\documentclass[12pt, letter]{article}
\usepackage{amssymb}
\usepackage{amsmath}
\usepackage{commath}
\usepackage{mathtools}
\usepackage{IEEEtrantools}

\usepackage{import}
\usepackage{paralist}
\usepackage{svg}
\usepackage{textcomp}
\usepackage{gensymb}
\usepackage{dsfont}
\usepackage{subfig}
\usepackage{graphics}
\usepackage{epsfig}
\usepackage[normalem]{ulem}
\usepackage[binary-units]{siunitx}
\usepackage{bm}
\usepackage[shortlabels]{enumitem}
\usepackage{upgreek}
\usepackage{todonotes}
\usepackage{scicite}
\usepackage{times}
\usepackage{xurl}
\usepackage{multirow}
\usepackage{makecell}
\usepackage{tabularx}
\usepackage{booktabs}
\usepackage[percent]{overpic}
\usepackage{xcolor}
\usepackage{transparent}
\usepackage{tikzscale}
\usepackage[htt]{hyphenat}
\usepackage{nameref}
\usepackage{chngcntr}
\usepackage{titling}
\usepackage{hyperref}

\DeclareMathOperator*{\minimize}{minimize}

\DeclareMathOperator*{\st}{subject~to}
\newcommand{\SO}{\mathrm{SO}}

\newcommand{\SE}{{\mathrm{SE}}}

\newcommand{\tm}{\textsuperscript{\texttrademark}~}

\renewcommand{\emph}{}

\newcommand\blfootnote[1]{%
    \bgroup
    \renewcommand\thefootnote{\fnsymbol{footnote}}%
    \renewcommand\thempfootnote{\fnsymbol{mpfootnote}}%
    \footnotetext[0]{#1}%
    \egroup
}

\newenvironment{sciabstract}{%
\begin{quote} \bf}
{\end{quote}}

\date{}

\begin{document}

\title{iCub3 Avatar System: \\
Enabling Remote Fully-Immersive\\
Embodiment of Humanoid Robots
}

\author{Stefano	Dafarra$^{*,1}$, Ugo Pattacini$^{2}$, Giulio Romualdi$^{1}$, Lorenzo Rapetti$^{1}$,\\
        Riccardo Grieco$^{1}$,  Kourosh Darvish$^{1,6}$, Gianluca Milani$^{1}$, Enrico Valli$^{1}$, \\
        Ines Sorrentino$^{1,3}$, Paolo Maria Viceconte$^{1,4}$, \\
        Alessandro Scalzo$^{2}$, Silvio Traversaro$^{1}$, \\
        Carlotta Sartore$^{1,3}$, Mohamed Elobaid$^{1}$, Nuno Guedelha$^{1}$, Connor Herron$^{5}$,\\
        Alexander Leonessa$^{5}$, Francesco Draicchio$^{7}$, Giorgio Metta$^{2}$,\\
        Marco Maggiali$^{2}$, Daniele Pucci$^{1,3}$\\
	    \\
\normalsize{$^{1}$ Artificial and Mechanical Intelligence, Istituto Italiano di Tecnologia, Genoa, Italy,}\\
\normalsize{$^{2}$ iCub Tech Facility, Istituto Italiano di Tecnologia, Genoa, Italy}\\
\normalsize{$^{3}$ School of Computer Science, University of Manchester, Manchester, UK}\\
\normalsize{$^{4}$ DIAG, Sapienza Università di Roma, Rome, Italy}\\
\normalsize{$^{5}$ Department of Mechanical Engineering, Virginia Tech, Blacksburg, VA, USA}\\
\normalsize{$^{6}$ Computer Science and Robotics Institute, University of Toronto, Toronto, Canada}\\
\normalsize{$^{7}$ DiMEILA, Istituto Nazionale Assicurazione Infortuni sul Lavoro (INAIL), Rome, Italy}\\
\normalsize{(Corresponding author contact: {\tt\small stefano.dafarra@iit.it})}
}

\baselineskip24pt

\maketitle

\begin{sciabstract}
We present an avatar system designed to facilitate the embodiment of humanoid robots by human operators, validated through iCub3, a humanoid  developed at the {Istituto Italiano di Tecnologia} (IIT){.}
{More precisely, t}he contribution of the paper is twofold: first, we present the humanoid iCub3 
{as a} robotic avatar which integrates the latest significant improvements after about fifteen years of development {of the iCub series}; 
second, we present a versatile {avatar} system 
enabling humans to embody humanoid robots encompassing aspects such as locomotion, manipulation, voice, and face expressions with comprehensive sensory feedback including visual, auditory, haptic, weight, and touch modalities. 
We {validate the system} by implementing {several} avatar {architecture} instances,
each tailored to specific requirements.
First, we  evaluated the optimized architecture for verbal, non-verbal, and physical interactions with a remote recipient. This testing involved the operator in Genoa and the avatar in the Biennale di Venezia, Venice -- about {290} Km away -- thus allowing the operator to visit remotely 
{the} Italian art exhibition.
Second, we evaluated the {optimised} architecture for recipient physical collaboration and public engagement on-stage, live, {at} the \emph{We Make Future} show, 
{a} prominent world digital innovation festival. In this instance, the operator was situated in Genoa while the avatar operates in Rimini -- about 300 Km away -- interacting with a recipient who entrusted the avatar a payload to carry on stage before an audience of approximately 2000 spectators. 
Third, we present the architecture implemented by the iCub Team for the ANA Avatar {XPrize} competition.

\end{sciabstract}

\paragraph{Summary:} We present an avatar system to embody the humanoid robot iCub3 for remote verbal, non-verbal and physical interaction.

\section*{Introduction}
The emergence of biological disasters and the evolution of digital virtual ecosystems necessitate the advancement of avatar technologies, enabling humans to inhabit either remote real locations or immersive virtual realities {(VR)}. The COVID-19 pandemic, for instance, highlighted the immature state of avatar technologies to facilitate effective human operations in distant locations~\cite{leidner2021covid}. Analogously, the renewed interest of the {e}ngineering community in {VR} systems is also fueled by the increasing applications of digital and virtual ecosystems across various sectors~\cite{Pillai2022XR}. The renewed impetus is underscored by initiatives like the ANA Avatar {XPrize}, a 10M\$ competition\cite{xprize_link} dedicated to creating avatar systems capable of transporting human presence to a remote real location in real-time. 
This paper 
contributes to the development of technologies and methods that empower humans to embody physical humanoid robot avatars for real-time operations in remote locations.

When attempting at creating physical avatars, one is tempted to apply the state of the art on \emph{telexistence} \cite{tachi2015telexistence}{.}
A telexistence system allows transferring, and possibly augmenting, the skills of the human operator to a robotic avatar. Intuitiveness is a key feature of the system, trading off the autonomy of the robotic avatar with the capabilities of the human operator to cope with unforeseen circumstances. Through the system, the operator is connected to the remote location while interacting with the environment or engaging with another person. {Cybernetic avatars can also have effects at the societal level, allowing people to contribute to society without constraints \cite{ishiguro2021realisation,horikawa2023cybernetic}.} 

{P}hysical avatar technologies benefit from the state of the art in telexistence.
A physical avatar system is mainly composed of three components that are often the constituents of telexistence systems: the \emph{physical avatar}, often a robot capable of navigating the environment; the \emph{operator system}, which is in charge of retargeting and tele-perception; the \emph{communication layer}, enabling communications between the avatar and the operator system. 

\emph{Physical avatars}
are often implemented with robots having locomotion capabilities. Typical solutions include multi-legged or wheeled robots~\cite{klamt2020remote,LENZ2023bimanual,schwarz2021nimbro,lentini2019alter}. In contexts where remote interaction with humans is crucial, humanoid robot avatars show great potential for existing and future applications{. T}he robot's human-likeness feature increases its acceptability, its social interaction performances, and the clarity of its intentions \cite{dragan2013legibility}. Also, when compared to wheeled or multi-legged robots, {a} bipedal system design can perform more complex movements in reduced and confined spaces. Humanoid robots thus represent an optimal starting point for a platform to embody humans in terms of locomotion, manipulation, verbal, and non-verbal interaction, allowing an operator to have direct control over the whole body of the robot \cite{darvish2023teleoperation,darvish2019whole,penco2019multimode}. The bipedal humanoid design, however, poses additional challenges due to the inherent instability of the robotic system. This complexity can be {handled} by letting the robot autonomously control its stability while achieving the desired tasks commanded by the remote operator, which may provide only high-level commands (for example, walking references) \cite{abi2018humanoid,elobaid2019telexistence,kim2013whole}.
In this case, the robot autonomously stabilizes the desired walking pattern and the lower-body motion of the robot is not synchronized with the operator's movements.

{Whilst navigating and manipulating the environment, the robot can sense its surroundings through specialized touch sensors. Touch feedback can have a noticeable effect on the manipulation capabilities of an avatar system \cite{fishel2020tactile}, but it can also enable teleoperated physical interaction \cite{kaplish2019motion}, giving rise to social implications.}
{In the context of \emph{social presence}, other avatar characteristics gain relevance, such as the control of facial expressions \cite{van2022comes, rochow2022vr}}.

\emph{The operator system}
often consists of a set of wearable devices and algorithms in charge of the so-called \emph{retargeting} and \emph{tele-perception} features \cite{luo2022towards}. The devices are often {VR} commercial products \cite{martinez2015telepresence,shin2022stereoscopic,vaz2022immersive}, or motion capture systems \cite{darvish2019whole,penco2019multimode}. 
To achieve bilateral feedback teleoperation, it is possible to leverage ad-hoc designed exoskeletons {\cite{LENZ2023bimanual, schwarm2019floating}}. 
However, exoskeletons can be very invasive, thus constraining the operator's motion in{,} sometimes{,} narrow envelopes.

\emph{The communication layer} connects the operator system to the physical avatar. It allows the different components of the {telexistence} system to communicate {with} each other in spite of potentially delayed networks \cite{schmaus2018preliminary}. The software suite that implements the communication layer is usually referred to as \emph{middleware}. Common middlewares are the \emph{Robot Operating System} (ROS) \cite{quigley2009ros}, and YARP \cite{fitzpatrick2014middle}.

{
\paragraph{}
Humanoid robots have been considered for a large variety of applications, ranging from rehabilitation of the elderly to interactions with autistic children \cite{tanioka2019nursing,alnajjar2020humanoid}. In many applications, humanoid robots are thought of as teleoperated machines, and social implications of this control mode received attention from the scientific community \cite{dave2022avatar}. These studies aim to afford the operator a sense of embodiment in the robot, reproducing solely the operator's movements, in an attempt to make the recipients engage with a small humanoid robot. Other applications focus on the retargeting of the operator's motions in full-size humanoids \cite{di2016multi}, but they do not consider the locomotion aspects.
In some cases, the operator can also be in charge of the balancing of the robot through lower-body \cite{ramos2018humanoid} or full-body exoskeletons \cite{ishiguro2018high,ishiguro2020bilateral}. In other cases, exoskeletons can provide haptic or vibrotactile feedback on the current balance status of the robot \cite{abi2018humanoid,brygo2014humanoid}. In contrast to the above efforts, we present in this paper a complete avatar system where the operator's motions are fully retargeted on the robot, including the locomotion intents. The robot is in charge of keeping its balance, while haptic and vibrotactile feedback is provided to the operator when the robot interacts with the environment.
}

{
Fully-fledged avatar systems often allow the operator to control the robot's manipulation and locomotion abilities while providing visual, auditory, and haptic feedback \cite{klamt2020remote}. A reduced teleoperation system has also been tested with the operator being an astronaut on the International Space Station \cite{lii2022introduction}. In the vast majority of the cases, the robot is either wheeled or in a sitting configuration \cite{tachi2015telexistence}. In this paper, we present a complete avatar system exploiting a humanoid robot, while also considering emotional aspects. In fact, we let the operator control the robot's facial expressions too, while receiving haptic feedback when the robot is touched. Moreover, we test the system with the operator and the robot positioned hundreds of kilometers apart.
}

{
The ANA Avatar XPrize competition provided a testing ground for avatar systems \cite{hauser2023xprize}. Most teams adopted a wheeled configuration \cite{schwarz2021nimbro,lentini2019alter,van2022comes}, or a hybrid legged-wheeled solution \cite{vaz2022immersive}. Concerning the equipment enabling the robot teleoperation (referred to as ``operator equipment''), both light commercial VR devices \cite{lentini2019alter,vaz2022immersive} or custom-made exoskeletons \cite{schwarz2021nimbro,van2022comes,luo2022towards} have been adopted. Our avatar system was the only one to complete tasks in the final stage utilizing bipedal locomotion with a lightweight set of operator devices, comprising both commercial and custom-made wearables.
}

\paragraph{}
The contribution of the paper is twofold.
First, we present the iCub3 humanoid robot. After about fifteen years of development of the iCub platform, the iCub3 is the latest iCub version with increased body size, optimized for locomotion and physical interaction tasks.
   
Second, we present a generic avatar system that allows an operator to embody humanoid robots.
The operator is given a set of lightweight and non-invasive devices, {comprising} iFeel, custom-made wearable technologies for motion and force tracking, developed by the Istituto Italiano di Tecnologia (IIT). The avatar system simultaneously transports the operator's locomotion, manipulation, voice, and, facial expressions to the avatar with visual, auditory, and haptic feedback (weight, touch).

We validate the avatar system using the iCub3 as {an} avatar. The validations consisted of four implementations of the avatar system, each designed to meet different objectives. 
First, a remote visit of the \emph{Italian Pavilion} within the \emph{Biennale dell'Architettura di Venezia}\cite{biennale_link}, where the operator was in Genoa and the avatar in Venice, at about \SI{290}{\km} distance. The objective here was to test the embodiment and the remote physical interaction via the iCub3 avatar system. The demonstration is visible in the Supplementary Movie and online in a longer format\cite{biennale_video}.
Second, a remote participation to the live show \emph{We Make Future}\cite{wmf_link}, where the operator was in Genoa and the avatar in Rimini, at about \SI{300}{\km} distance, in front of 2000 spectators. The objective here was to perform a physical collaboration task with a remote recipient while entertaining an audience. The full show is available online\cite{wmf_video}, while the iCub3 demonstration is presented also in the Supplementary Movie.
Third, the \emph{ANA Avatar {XPrize} Semifinals}, where the objective of the implemented architecture was to maximize the sense of presence and shared situational awareness with a recipient while having precise control of the robot end effectors.
Fourth, the \emph{ANA Avatar {XPrize} Finals}, where the avatar system was focused on complex locomotion and {manipulation} of heavy and textured objects.

\section*{Results}

This section introduces the validation results of the generic avatar system architecture that will be presented in the Methods. In particular, the validation scenarios listed in the Contribution paragraph of the Introduction define different requirements and shape different objectives -- see  Table \ref{tab:validation_requirements}. 

\subsection*{Remote teleoperation: iCub3 explores the Biennale di Venezia}

The iCub3 avatar system underwent testing in a demonstration where the operator was situated at the IIT {premises} in {Genoa}, Italy, and the iCub3 robot was positioned in the \emph{Italian Pavilion} of the \emph{Biennale dell'Architettura di Venezia} located in Venice, Italy. Consequently, the operator and the robot were approximately \SI{290}{\kilo\metre} apart, ``linked'' through a standard fiber optic internet {network}. The primary objective of this validation was to establish an architecture enabling the human operator to possess verbal, non-verbal, and physical interaction capabilities with a person at the remote location, referred to here as {the} \emph{recipient}. This demonstration was made possible through a collaboration between IIT and the Italian Ministry of Culture, and the test was conducted on November the 8th, 2021.

At the time of the demonstration, the logging systems described in the Methods were not available yet, hence we have no numerical data to present. As a matter of fact, this demonstration taught us the importance of such systems.
The latency introduced by the communication channel only has been constantly monitored, remaining stably below \SI{25}{\milli\second}. This reduced latency did not affect the operator experience. In addition, the delay did not hinder the robot's stability since its control system ensured balance independently from the network configurations.
A video of the demonstration is available as part of the Supplementary Movie. A more detailed version is available online\cite{biennale_video}, to which we refer in the following.

\begin{figure}[tpb]
    \centering
    \includegraphics[width=\columnwidth]{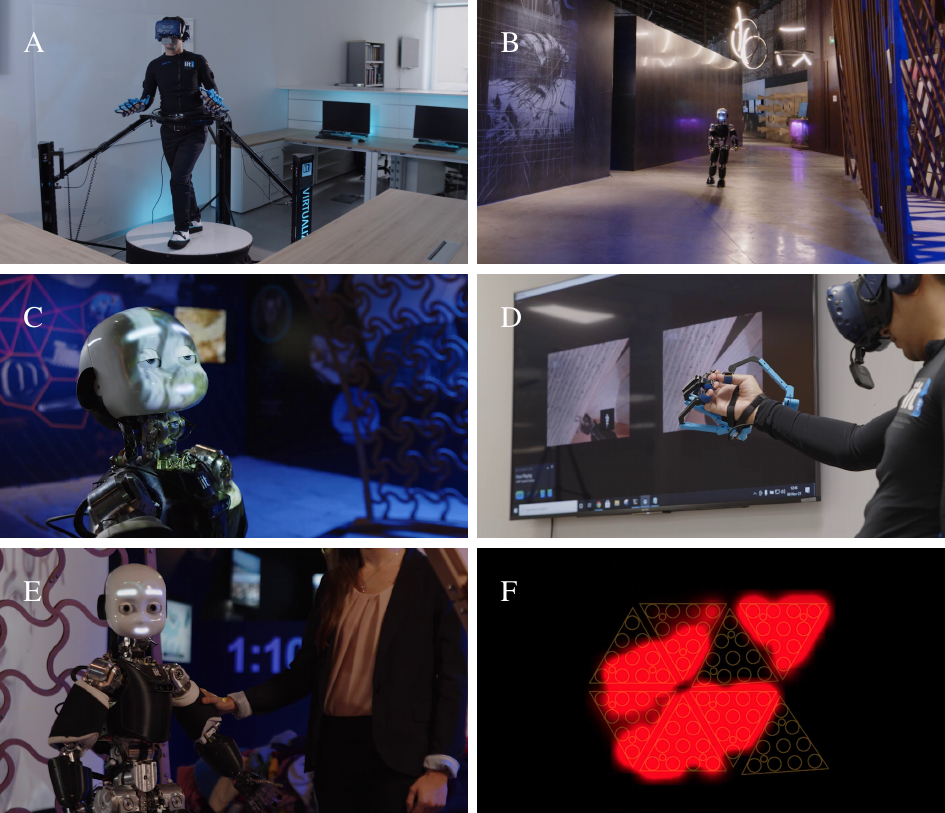}
    \caption{\textbf{iCub3 explores the Biennale di Venezia}. Snapshots of the video\cite{biennale_video} demonstrating the remote teleoperation of iCub3 at the Italian Pavillion of the Biennale di Venezia. The operator navigates the remote venue via iCub3 (\textbf{A}, \textbf{B}). The operator controls the iCub3 eyelids in response to bright light (\textbf{C}). The operator remotely grasps a piece of tissue through iCub3 (\textbf{D}). The robot is touched on the arm (\textbf{E}). The robot skin, whose activation is represented in (\textbf{F}), triggers the body haptic feedback on the operator.}
    \label{fig:iCub_venice}
\end{figure}

The first part of the video, up to time 0:55, is dedicated to the preparation of the operator, who wore the devices mentioned in the Methods section.
At time 1:25, and later at 1:51, the operator exploited the robot \emph{locomotion} capabilities. In particular, by walking inside the Cyberith Virtualizer\tm platform, the operator was able to walk around the venue, as shown in Fig. \ref{fig:iCub_venice}(A and B). At 1:26, the operator then interacted through the avatar with the \emph{recipient}. In this context, the \emph{visual} and \emph{auditory feedback} were fundamental for a proficient verbal interaction. The \emph{face expressions retargeting}, demonstrated in Fig. \ref{fig:operator_Retargeting}(A and B), enabled the non-verbal interaction, allowing the operator to smile to the \emph{recipient}, or to close the eyes in case of bright light, as demonstrated in Fig. \ref{fig:iCub_venice}C, and in minute 2:19 of the detailed video.
At time 1:58 and 2:08 the operator exploited the control over the robot body to express body language and to point at some installations while interacting with the \emph{recipient}.
The \emph{touch feedback} was fundamental when the operator interacted with the venue at time 2:43, Fig. \ref{fig:iCub_venice}D. The \emph{manipulation} and fine control of each robot finger allowed the operator to touch the installation with delicacy while perceiving \emph{haptic feedback}.

Finally, at time 2:52, we showcased the importance of the body \emph{haptic feedback} for immersive interaction. As shown in Fig. \ref{fig:iCub_venice}(E and F), the \emph{recipient} reached the robot from outside its field of view. She then touched the robot's arm. The robot skin perceived the touch and triggered the body \emph{haptic feedback}. Hence, the operator \emph{perceived} the remote touch and turned toward the \emph{recipient} direction.
The remote visit ended with the operator and the \emph{recipient} sharing a hug, highlighting the emotional implications of such a rich interaction.

\subsection*{Remote Teleoperation: iCub3 on the stage of the ``We Make Future'' Festival}

\begin{figure}[tpb]
    \centering
    \includegraphics[width=0.85\columnwidth]{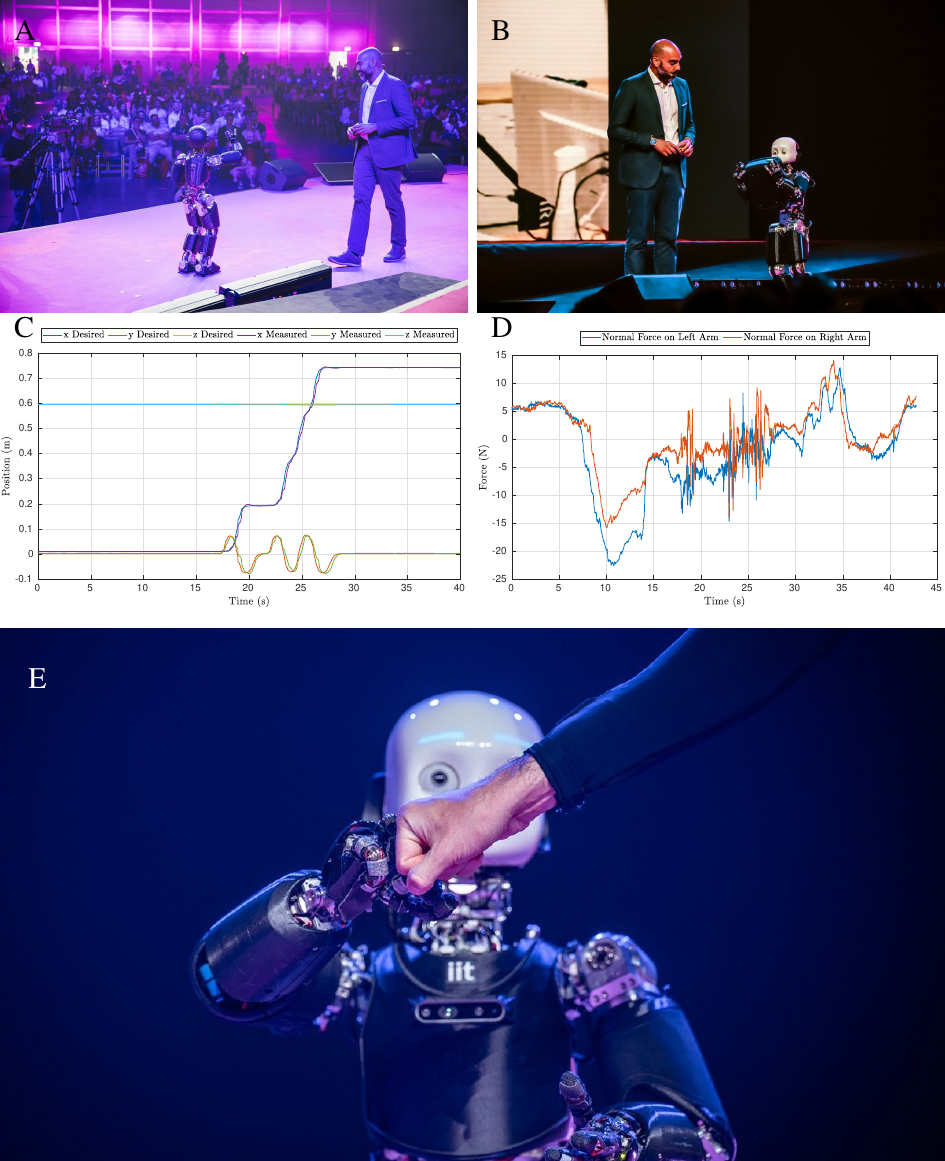}
    \caption{\textbf{iCub3 on the stage of the ``We Make Future'' Festival}. iCub3 is on the stage of a tech fair, while being teleoperated from remote\cite{wmf_video}. iCub3 interacts with the audience (\textbf{A}). iCub3 holding a box (\textbf{B}). Plot of the center of mass tracking of the robot while walking with a box (\textbf{C}). Plot of the vertical force measured on the robot arms while walking. A recipient hands the robot the box at around \SI{10}{\second}. Then, the box is taken back by the recipient after the robot stopped walking (\textbf{D}). {iCub3 interacts with the recipient (\textbf{E}).}}
    \label{fig:iCub3_wmf}
\end{figure}

On June 16th, 2022, iCub3 made a guest appearance on the stage of the ``We Make Future'' Festival \cite{wmf_link} in Rimini, as depicted in Fig. \ref{fig:iCub3_wmf}(A). The robot was teleoperated from {Genoa}, situated approximately \SI{300}{\kilo\metre} from the venue in Rimini, { with a network delay comparable to that described in the previous subsection}. The demonstration, featured in the Supplementary Movie and accessible online\cite{wmf_video}, aimed to validate an architecture that provided the operator with verbal, non-verbal{,} and physical interaction capabilities with {another} person in the remote location. Additionally, the avatar was tasked with engaging the public, as illustrated in Fig. \ref{fig:iCub3_wmf}(E).

The implemented instance of the avatar system was similar to the one adopted for the remote visit at the Biennale di Venezia, except for the use of the VIVE\tm trackers in conjunction with the iFeel suit for improved Cartesian control of the robot's hands. Moreover, the iFeel haptic nodes were used to inform the operator about the weight held by the robot. In fact, during the demo, the robot was supposed to walk while carrying a weight, see Fig. \ref{fig:iCub3_wmf}(B). Figure \ref{fig:iCub3_wmf}(C) shows the center of mass tracking while walking with a box weighing about \SI{0.5}{\kilo\gram}. The controller was unaware of the additional weight and it considered it an external disturbance. 
{While walking, the robot controller favored the tracking of the walking-related trajectories compared to the retargeting trajectories. Thus, in case of conflicts, the robot balance was preserved at the expense of the retargeting performances. More details are in the Methods section.}
The external force induced by the weight of the box was measured by the force-torque (F/T) sensors installed on the robot arms. Figure \ref{fig:iCub3_wmf}(D) displays the vertical component of the measured forces exerted on the robot arms. At $t=$~\SI{0}{\second}, we can notice an initial offset measured by the F/Ts. Then, around $t=$~\SI{10}{\second}, the box was handed to the robot. At about $t=$~\SI{17}{\second}, the robot started walking and the impacts with the ground induced some disturbances in the measured linear force. Finally, after the robot stopped walking, at $t=$~\SI{27}{\second} the recipient took the box back from the robot.

\subsection*{ANA Avatar XPrize Semifinals}

In the semifinals of the ANA Avatar XPrize competition, the iCub3 Avatar system underwent a series of tasks while being operated by two competition judges. Following a half-hour training session, each judge was required to execute three scenarios, each consisting of six atomic tasks. The scenarios were designed to assess the system's capabilities, encompassing visual and auditory perception, gaze control, gestures, haptics capabilities, manipulation, grasping, and mobility. The team's overall score during the semifinals hinged on the proficiency exhibited in these aspects. Additional judging factors included the quality of interaction between the operator and another human being, the recipient. Hence, body language, emotional expression, and shared situational awareness were also taken into consideration. Moreover, being the system teleoperated by a novice operator, the intuitiveness and the ease of use of the teleoperation system were implicitly factored into the evaluation. Some tasks were replicated among different scenarios, while others specifically necessitated the operator to communicate to the recipient solely through the avatar. In the following, we present a meaningful excerpt of the semifinal tasks, defining how the iCub3 Avatar system was used to approach them.

In total, 38 teams from 16 countries qualified for the semifinal stage\cite{seminfinals_teams}. The total score assigned to the iCub3 Avatar system was 95 over a total of 100 points {\cite{hauser2023xprize}}, which was worth second place overall\cite{semifinals_score}. The semifinals took place on the 21st of March 2022, with the XPrize judges visiting the IIT labs in {Genoa}.

\subsubsection*{Puzzle task}
\begin{figure}[tpb]
    \centering
    \includegraphics[width=0.8\columnwidth]{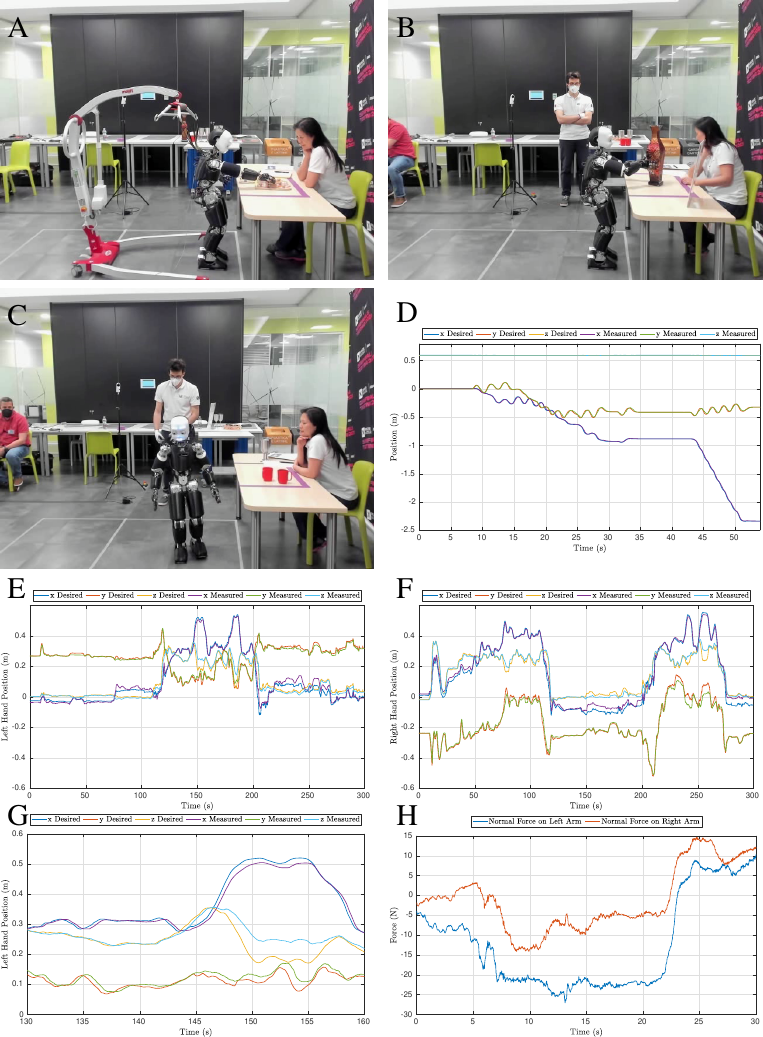}

    \caption{\textbf{iCub3 at the XPrize semifinals}. Pictures and plots of the iCub3 avatar system performance at the Xprize semifinals. iCub3 manipulating one puzzle piece (\textbf{A}). iCub3 checking the texture of the vase (\textbf{B}). iCub3 walking during the XPrize semifinals (\textbf{C}).CoM tracking during the walking task (\textbf{D}). The left and right hand Cartesian errors during the puzzle task (\textbf{E}) and (\textbf{F}). Magnified version of (\textbf{E}) to highlight the Cartesian tracking and lags (\textbf{G}). Normal force measured by the hands while lifting the vases and putting it back in place (\textbf{H}).}
    \label{fig:semifinals}
\end{figure}

In order to test the manipulation and grasping capabilities of the system, one of the semifinal tasks required the operator to collaborate with the recipient, via the iCub3 avatar, on a toddler-like puzzle, shown in Fig. {\ref{fig:semifinals_puzzle}}. This task required high accuracy in the placement of the robot hand. To improve the cartesian tracking of the operator's hand movements, we resorted to the VIVE\tm trackers in conjunction with the iFeel nodes. At the same time, the fine control of the robot fingers allowed the operator to firmly grasp and place the puzzle pieces, Fig. \ref{fig:semifinals}({A}). Figures \ref{fig:semifinals}(E and F) plot the cartesian error for the left and right hand. In particular, the desired position has been reconstructed from the joint values obtained from the \emph{manipulation interface} described in the Methods, whereas the measured position is reconstructed from the joint values measured on the robot. Both quantities are expressed with respect to a frame attached to the robot pelvis link, with the z-axis pointing upward, and the x-axis pointing forward. 
The measured position largely follows the desired position, with some exceptions. For example, at around $t=$~\SI{255}{\second}, a small offset is visible on the $z$ and $y$ axes. This offset can be blamed on the balancing controller presented in the Methods. In fact, wide motions of the arms and torso can affect the CoM position, causing the balancing controller to intervene and potentially reduce the Cartesian tracking performances.
{Figure \ref{fig:semifinals}(G) presents a magnified version of the left hand Cartesian tracking. There is an error of about \SI{5}{\centi\meter} in the $z$ direction. The operator was requesting the robot hand to be in a lower position, but this relatively large error may indicate that the robot hand was already touching the tabletop. From the same figure, there is a retargeting lag in the order of \SI{0.5}{\second}.}
{The Cartesian tracking does not necessarily indicate the performance of the system in completing the task, but demonstrates that the motion of the operator was tracked on the robot.}
{Moreover, for this specific task, the visual feedback was the most useful for the operator, who was able to compensate for lag and tracking errors by looking directly at the robot hand and performing movements at low speed.}

\subsubsection*{Weight task}
Another semifinal task consisted in the operator detecting the weight of a vase through the avatar. The vase was placed on a table in front of the robot. Due to the limited dimension of the iCub hands, we instructed the operator to use both robot hands to determine the weight of the object. Similarly to the remote teleoperation experiment at the ``We Make Future Festival'', we exploited the iFeel haptic nodes to provide the operator an indication of the weight of the vase via haptic feedback. Moreover, we displayed in the headset the numeric value of the weight estimated via the arms' F/T sensors. Figure \ref{fig:semifinals}{(H)} displays the normal force estimated by both arms when lifting the vase and putting it back in place. The weight of the vase was about \SI{1}{\kilo\gram}, and the F/Ts partially overestimated it. The task was performed using two hands, thus introducing some internal forces that were measured by the F/Ts. When placing the vase on the table, the robot also gently nudged it against the tabletop, resulting in a positive force measured by the F/Ts.

\subsubsection*{Texture task}
{T}he avatar was supposed to {let} the operator feel the texture embossed on the surface of the same vase of the previous paragraph. In Fig. \ref{fig:semifinals}({B}), the operator was keeping the vase still with the robot's left hand, while he scanned the surface with the right fingers to detect the embossed texture.
For this task, we exploited the sensorized skin installed on the robot's fingers. The activation of the sensing elements triggered a vibration on the corresponding operator's finger via the haptic glove. The mapping function was tailored to be sensitive to light touches, without being too distractive for power grasps. The activation of the skin and the consequent vibration feedback was not part of the logged data and{,} unfortunately, we have no numerical data to show for this task.

\subsubsection*{Locomotion task}

The Semifinals tested the {system} locomotion capabilities, Fig. \ref{fig:semifinals}({C}). The robot {had to} move away from the table and walk a couple of meters to reach a designated area, indicated by tape on the ground.
We exploited the Cyberith Virtualizer to trigger the robot's motion. In particular, we instructed the operator to first turn around. At this point, the robot autonomously defined a set of steps to turn in place without moving forward (thus avoiding the table). Once fully turned, the operator started walking forward inside the Virtualizer to reach the designated area.

Figure \ref{fig:semifinals}({D}) shows the CoM tracking of the balancing controller described in the Methods, while walking {away from} the table and to the goal position.

\subsection*{ANA Avatar XPrize Finals}

We used the iCub3 Avatar System in the ANA Avatar XPrize finals{,} conducted at the Long Beach Convention Center in Los Angeles, California, on 1-5 November 2022. This conclusive phase of the competition featured the participation of 17 teams. An overview of the competition test course is illustrated in Fig. \ref{fig:finals_stage}. 

In contrast to the semifinals, the competition's focus shifted, emphasizing heavy-duty tasks over the interaction between the avatar and the recipient. Similar to the semifinals, the operator maneuvering the avatar system was an XPrize judge. However, dressing and training time were constrained to a total of 45 minutes. The Avatar system was tested in a single scenario themed on the exploration of another planet. The tasks are summarized in Table \ref{tab:finals_tasks}.

Points were awarded to the Avatar system upon the completion of each task, with the requirement to accomplish all tasks within a total time of {less} than 25 minutes. Failure to complete a single task resulted in the termination of the trial. During our scored trial, the robot collided with one of the door's pillars, leading to a fall that precluded further participation in the competition. {W}e ranked 14th {\cite{hauser2023xprize}}. The video of the trial is included in the Supplementary Video and is available online\cite{finals_video}. The following section outlines our approach in deploying the iCub3 avatar system for the various tasks, even those that were not successfully completed during the competition.

One additional complication was the Wi-Fi connectivity. The wireless connection to the robot was provided by the organizers and the maximum bandwidth provided at the edges of the competition course was below \SI{100}{\mega\byte\per\sec}{, whereas it reached \SI{150}{\mega\byte\per\sec} toward the center of the course}. 
{We minimized the network usage to avoid delays in the visual-manipulation pipeline and we were unable to record logging data during the competition.}
Hence, we lack numerical data on the tasks performed during the finals.

\subsubsection*{Switch task}
The switch used during the XPrize Finals was a widely available commercial product. The handle required about \SI{30}{\newton} to be moved. Such force could have had destructive effects on the robot hand/wrist mechanism, which was designated to hold light objects. Hence, we equipped the robot with a small plastic cylinder installed directly on the forearm assembly. On the site, the internal handle resistance had been almost completely removed by the organizers. Nonetheless, the Operator exploited the cylinder to activate the switch, Fig. \ref{fig:iCub3_finals}B.

\subsubsection*{Locomotion tasks}
Compared to the semifinals, the Avatar locomotion had significance during the XPrize finals. Due to the time necessary to set up the Cyberith Virtualizer\tm, and the necessity of walking sideways, we adopted the iFeel walking solution, described in the Methods.

\begin{figure}[tpb]
    \centering
    \includegraphics[width=\columnwidth]{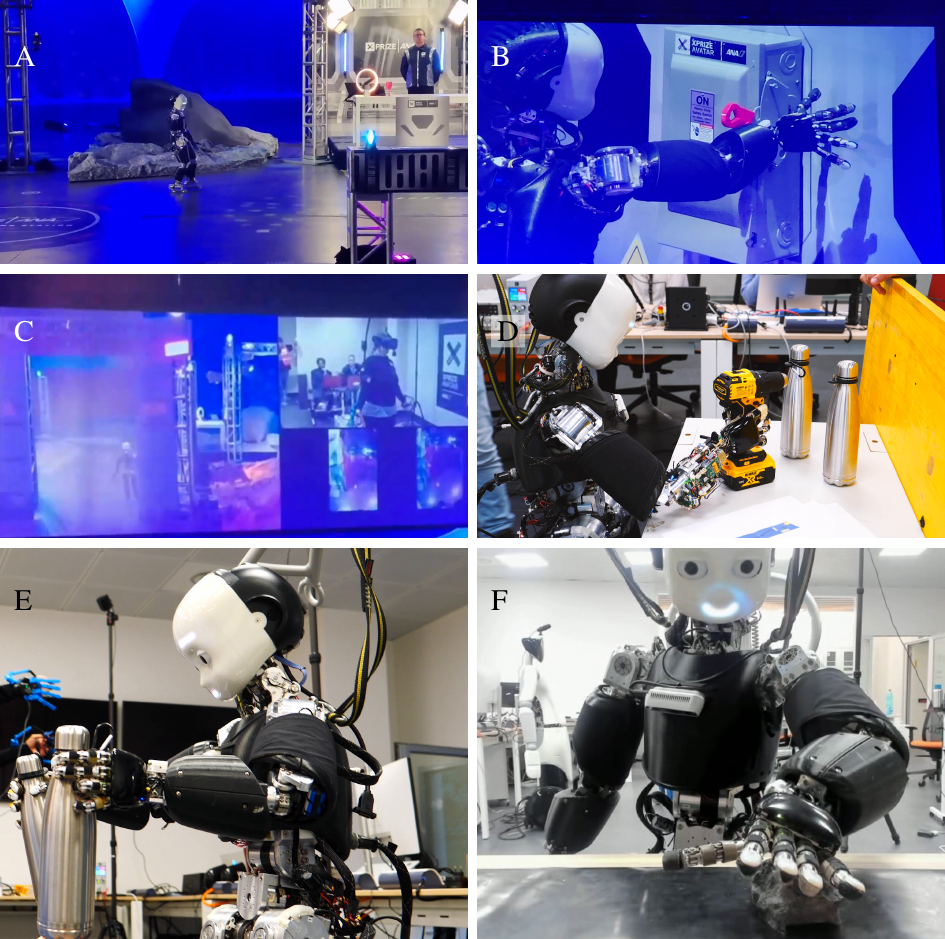}
    \caption{\textbf{iCub3 at the XPrize finals}. Pictures of the iCub3 avatar system performance at the XPrize finals. The first three pictures have been shot by the authors while on the finals course. The last three pictures have been taken during tests on the lab prior to the finals. iCub3 walking on the course, themed on the exploration of another planet (\textbf{A}). iCub3 activating the switch using a plastic cylinder installed on the wrist (\textbf{B}). A portion of a video of the iCub3 finals trial, while hitting the door. The Operator's view is visible in the bottom right corner  (\textbf{C}). iCub3 {activating the drill} (\textbf{D}). iCub3 {holding the XPrize finals canisters} (\textbf{E}). iCub3 making contact with a rough textured rock via the hand palm skin (\textbf{F}).}
    \label{fig:iCub3_finals}
\end{figure}

To improve the robustness of the walking motion, we increased the walking controller frequency to \SI{500}{\hertz}. iCub3 was the only robot in the finals successfully exploiting bipedal locomotion, Fig. \ref{fig:iCub3_finals}(A). Whilst trying to pass through the door, the operator underestimated the dimensions of the robot, passing excessively close to one of the pillars. The robot arms were controlled with a rigid position controller (to allow fine control while manipulating), and when one arm hit the pillar, the resulting reaction force destabilized the robot causing the fall, Fig. \ref{fig:iCub3_finals}(C), thus ending our trial. The next subsections present our approach to the tasks we were not able to complete during the scored trial, with insights from the tests prior to the finals.

\subsubsection*{Bottles task}

The estimation of the heavy canister exploited the same infrastructure of the weight task of the semifinals. On the other hand, compared to the previous case, we tested a configuration where the robot did not hold the object with two hands, but with a single one. Moreover, the weight estimation coming from the arms was printed separately on the headset. In this way, the Operator could have easily compared the weight of two canisters while holding them in hand like in Fig. \ref{fig:iCub3_finals}(E).

\subsubsection*{Drill task}
The grasping and activation of the drill would have been a difficult task for the iCub3 wrist/hand mechanism. The main problems were the weight of the drill, about \SI{2.5}{\kilo\gram}, and the force necessary to activate the trigger, about \SI{15}{\newton}.
The iCub wrist was not strong enough to fully sustain the drill weight. Nonetheless, we noticed that when trying to raise it, one of the wrist joint would have reached its mechanical limit. As a consequence, the weight of the tool was sustained by the forearm at the cost of reduced control over the orientation of the tool.
At the same time, the iCub3 index alone was not strong enough to pull the trigger. To circumvent this issue, we installed a new gearbox on both the index and middle finger motors. The new gearbox had a reduction ratio four times higher. Moreover, the index finger appeared to be too short to operate the trigger successfully. As a consequence, we replaced the index finger with another middle finger, which was \SI{5}{\milli\meter} longer. Finally, we tied the index and middle fingers together to use them jointly. Figure \ref{fig:iCub3_finals}(D) shows iCub3 activating the drill.

\subsubsection*{Texture task}

The texture task required to identify a rough textured rock. To this end, we took advantage of the artificial skin covering the robot hand palms. Since the rocks were light and not fastened to the table, they could have easily slipped away.
Therefore our approach was to make contact with the rocks from the top using the sensorized palm as shown in Fig. \ref{fig:iCub3_finals}(F). 

When contact was detected, a vibration pattern resembling either plain or rough texture was triggered on the Operator's hand.
For the selection of the vibration pattern, we relied on a neural network trained to classify the type of contact (rough or plain) from the tactile sensors' activations. 
In particular, each sensorized palm included 48 tactile sensors providing measurements in the numeric range [0,255]. The higher the value, the higher the measured pressure. 
We interpreted these measurements as a 9x11-pixels grayscale image, where each pixel, excluding padding, corresponded to a tactile sensor.
Fig. \ref{fig:texture_task_details}(A and B) shows sample images retrieved from the palm in contact with a plain and a rough rock, respectively. 
Such images represent the input for our binary classifier, a customized version of the well-known AlexNet architecture~\cite{krizhevsky2012imagenet} scaled in size
by a factor of 32 to meet real-time inference constraints and
equipped with smaller convolutional filters and less max-pooling layers to cope with our low-dimensional input.
We trained our classifier for 25 epochs on a training dataset consisting of around 150 contacts per class, using batches of 32 samples and the Adam optimizer~\cite{kingma2015adam}.
On our test dataset which included around 40 contacts, the overall trained model accuracy was 78\% - see Fig. \ref{fig:texture_task_details}(C).

\section*{Discussion}

We present a set of validations where an operator teleoperated the humanoid robot iCub3 to visit a remote exhibition, or performed a live exhibition on a stage. 
The operator was able to walk while interacting physically, verbally and non-verbally with a recipient through the  avatar. We also demonstrate the iCub3 avatar system capabilities by participating to the ANA Avatar XPrize international competition. In this context, the system proved to be very immersive and easy to use, given the placement at the semifinals. In the following, we provide a series of insights and design recommendations. {Nonetheless, the XPrize finals proved to be a severe testing ground for our system, allowing us to identify a series of shortcomings.}

{
\subsection*{Design recommendations for system usability and insights}
In this section, we outline the key takeaways from the design process of the iCub3 avatar system. These lessons concern our context and the specific challenges we encountered.
We acknowledge that avatar systems are inherently diverse, and what worked well in our scenario may not generalize to other applications. 
Nonetheless, we believe our lessons contribute valuable experiential knowledge to the field, potentially useful to many researchers working on humanoid robot avatars.
}

{
\paragraph{Tradeoff between the operator's physical effort and the transparency}
Operator movements can be mapped seamlessly onto the robot. Nonetheless, operators may need to put effort in order to move their body against gravity, or to maintain balance. All the more, if the operator has to move to trigger the robot locomotion, the operator's energy expenditure may not be sustainable if the robot has to walk for long distances. The use of supportive devices and equipment (like lightweight lower-body exoskeletons, chairs, or similar) can circumvent this issue at the expense of the embodiment. In fact, these devices also limit the operator's motion, constraining their sense of presence in the remote location. 
In brief, light and wearable devices together with the possibility of moving freely provide high transparency and immersion, at the cost of higher physical stress for the operator.
}

{
\paragraph{Avatar design}
Humanoid robots are often considered as ``general purpose'', meaning that their human likeness can be useful in an unstructured environment at a human scale. In contrast, environments characterized by wide spaces and smooth flat ground shall encourage the usage of wheeled robots against those implementing legged locomotion. 
At the same time, in case of narrow spaces or irregular terrain, legged locomotion should be exploited by legged robots. In this respect, the operator should have the possibility to define with more precision where the feet need to be placed. In our case, a possible approach could be to extend the iFeel walking solution presented in the ``Manipulation interfaces'' section. For example, a particular operator's movement with one foot may be interpreted as a ``manual mode'' trigger, enabling direct control over the corresponding foot position.
The robustness required to operate in a given environment is another element to consider. In the case of a humanoid robot, it is necessary to consider the possibility of a fall, thus implementing strategies that can reduce the resulting effects. Moreover, the robot should have some degree of autonomy to keep the balance while adapting to the environment.}

{From the acceptability point of view, iCub3 appeared to have high scores when engaging with the recipient mostly because of its humanoid shape with relatively small dimensions, and its physical resemblance to a child. This qualitative observation is supported by recent studies showing that robots with faces able to follow the recipient's gaze increase their likeability  \cite{willemse2018robot}. However, the use of a robotic head does not allow the recipient to immediately recognize the operator, which may impair the overall goal of making a robot avatar. Similarly, the use of robotic hands with anthropomorphic sizes increases the robot's human likeness. However, the resulting mechanical complexity may limit the applicability to tasks due to the maximum force exerted by each finger, as it was for our iCub3 where hands could only support relatively light external perturbations.}
{In brief, human-like features seem to increase acceptability and engagement for the recipient but might be a limiting factor in case of heavy-duty tasks in simple environments.}

{Compared to its previous versions, iCub3 proved to be a much more robust robot. In particular, the absence of tendons in the legs and shoulders consistently reduced the maintenance time, since tendons tend to break over time. Moreover, iCub3 calibrates its joint position sensors at every startup by moving each joint to the hard stops. This ensures the repeatability of the robot's motions.}

{
\paragraph{Tradeoff between modularity and ease of use}
Having a degree of modularity at the software level helps in developing and integrating different technologies into the robot. 
This allows one to enable and disable features, thus having a teleoperation system that meets multiple requirements. At the same time, there could be many operational units running in parallel, each one fallible in different ways. This might increase the complexity of having everything up and running. On the contrary, a monolithic system with a single ``on-off'' button, where everything is interconnected, might be easier to start, but more difficult to recover in case of failures in one of its subsystems. The initial component development should be as separate as possible, working then on orchestration tools to start all the different parts in the correct order. A second layer to automatically recover in case of failure can render the system more robust.
}

{At the hardware level, we can extend the modularity concept in terms of acceptability as well. Not all possible operators might feel comfortable with a given wearable device. In more extreme cases like people with disabilities, some devices might not be used at all. Therefore, the flexibility of the types of operator devices allows for accommodating a wider range of potential users. }

{
\paragraph{Tradeoff between off-the-shelf technology and in-house development}
When designing the avatar system, we advocate the good engineering practice of exploiting existing technologies, thus limiting the integration cost to the development of the layers to establish the connection with the existing architecture. This cost is often proportional to the flexibility of the architecture, as mentioned in the point above. In our system, this has been the case for the VR headsets, for example, where we employed commercial devices only. Nonetheless, in some cases, it has been proven useful to ``reinvent the wheel''. When a particular technology is aligned with one's research direction, an attempt to develop a similar technology from a fundamental level can be useful and insightful, although very time-consuming. The end result might allow large customizability and extensibility. As an example, the in-house development of FT sensors allowed us to integrate them into shoes, an application that stemmed from the initial robotic application. 
}

{
\paragraph{Use of agile for team management}
When dealing with the organization of demos and competitions, it is fundamental to organize the work of the different team components. We adopt an agile methodology, common in project management, but particularly shaped for robotics research. In particular, we divide the work into biweekly \emph{sprints}. Each week, the team members join in \emph{update meetings} to discuss the progress and eventual difficulties. When close to an important event, the frequency of the updates increases by implementing \emph{standup meetings}. The team components are encouraged to detail their progress in \texttt{GitHub} issues, thus providing implicit documentation. This proved to be fundamental to get prepared for the events detailed in the Results section.
}

\subsection*{Shortcomings identified during the XPrize finals}
In the spirit of full transparency and continuous improvement, it is essential to acknowledge and discuss the limitations of our system. We believe that identifying and understanding these shortcomings contributes to the portrayal of our work and can serve as a foundation for possible enhancements. In the following sections, we delineate key areas where our system exhibited room for improvement, thus identifying the challenges inherent in its current state.
{
\paragraph{Operator system}
The avatar system allows the user to have direct control of different behaviors of the avatar at the same time, thus requiring the operator to undergo a constant and heavy cognitive load. One key difficulty is related to the sense of depth and the estimation of the actual avatar occupancy in the space.
}

{
Controlling the robot walking by stepping in place seems to improve the immersivity of the system to the point where the operator starts wandering unintentionally. During a trial run of the XPrize finals, the operator also felt like he was losing his balance, especially when he was able to see his physical body through the robot cameras. Although this condition is rare, it raises some concerns related to the use of supporting equipment for the operators at the cost of some degree of immersivity.
}

{
Another point to consider is lag of the video stream. The camera feed represents the principal source of feedback used by the operator to control the position of the robot arms in space. The delay in the visual feedback leads the operator to be more cautious, thus requiring more time to perform a task, which in our experience also resulted in an increased cognitive load. When performing a fine manipulation task, the operator would often look at the robot hand through the vision system, and perform small adjustments accordingly. However, due to the lag in the vision system, the corresponding robot motion might ``overshoot'' the operator's intentions. This issue might be addressed by using multiple types of feedback. If the operator has to grasp an object, ``feeling'' the object in the hand before actually seeing it, might make the operator understand that the task has been accomplished. The haptic feedback can usually be acquired and sent at a higher frequency with less lag compared to the visual feedback, but, at the same time, this desynchronization between feedbacks increases the cognitive load.
}

{
\paragraph{Communication layer}
During the XPrize finals, our team was particularly affected by Wi-Fi issues. 
Apart from the natural interferences that occur in an event with a multitude of electronic devices and wireless networks, we identified that the WiFi antennas installed on the robot were too small. The effect of the network difficulties is also visible in the video of the scored trial\cite{finals_video} where the audio and the camera feed are strongly delayed and with numerous drops. The smoothness of the image feed then increases later in the course, since the wireless signal is stronger. This also indicates that using more complex compression techniques for audio and video is important. In our case, the audio is not compressed, whereas the camera images are compressed at a constant rate. Given that the signal strength was varying across the field, having a variable rate compression algorithm could have helped.
}

{
\paragraph{Avatar}
The use of a humanoid robot as an avatar poses many challenges. First of all, it is inherently ``unstable'', consequently any malfunction or unexpected disturbance might cause the robot to fall. In the iCub3 particular case, the amount of exteroceptive sensors is reduced to the cameras and the Intel Realsense\tm in the torso, limiting the possibility of the operator realizing if there are obstacles close to the robot. This might result in a problem in case of a crowded or delicate environment. Some degree of shared autonomy could help the operator avoid hitting obstacles in this context. At the same time, exploiting more compliance, and step recovery strategies could have helped us in recovery from the unexpected push.}

{The robot's motion can be considered non-continuous, dictated by the location of the footsteps. The operator cannot choose freely where to place the footsteps, whereas the robot has to necessarily alternate side motions while proceeding forward. Hence, the motion of the robot might appear unpredictable. Thus, an autonomous collision avoidance system could reduce the cognitive load required from the operator, but this would necessitate dedicated sensors like LIDARs. 
}

{
The hands represent another point of discussion. The iCub3 hands are a complex mechanical system designed to finely manipulate light objects. This characteristic is also a consequence of the small space available to place the motors to control all the fingers. Consequently, we had difficulties when faced with the task of utilizing a heavy object, like a drill. The complexity of the hand made it very difficult to apply last-minute changes, where a considerable amount of time is required to design and machine new parts. A more modular approach could have helped us fine-tune the hand according to the required tasks.
}

{Despite these issues, the XPrize finals allowed us to test our system to its limits, and we were the only team able to exploit bipedal locomotion to complete a task. The finals also pushed us {to exploit} more of the avatar technologies on the operator side. In fact, we use the same F/T sensors in the robot's feet, and on the operator's shoes. The operator and the robot technologies also share similarities at the code level. For example, the inverse kinematics approach is similar on both sides. Hence, the iCub3 avatar system represents an organic ensemble where the components are connected at a logical, hardware, and software level. }

\section*{Material and Methods}

The outcomes presented in the Results are obtained by implementing several instances of the  avatar system architecture detailed in this section. In particular, the avatar architecture is composed of two main interfaces, namely the \emph{teleoperation} and the \emph{teleperception} interface, whereas a  physical network establishes a logical link between the  components of these two logical interfaces -- see Fig. \ref{fig:teleoperation_system}. 

The \emph{teleoperation} interface is composed of two components, \emph{retargeting} and \emph{control}. The former collects the operator's actions, intentions and expressions via a set of devices the operator wears. These inputs are transmitted in the form of references to the avatar \emph{control}.

The second interface, the \emph{teleperception}, is composed of the \emph{measurements} and \emph{feedback} components. The \emph{measurements} retrieved by the robot are transmitted to the operator as a \emph{feedback}, providing a first-person perspective of the surroundings sensed by the robot.

\subsection*{The Avatar: iCub3}
The longstanding iCub platform has been evolving along several directions over the last fifteen years \cite{Nataleeaaq1026}. However, all its versions, which range from v1.0 to v2.9\cite{icub_versions}, are based on a humanoid robot having mostly the same morphology, size, joint topology, actuation, and transmission mechanisms. In other words, the evolution of iCub mechanics never focused on the robot height --~which was kept constant at about one meter~-- nor the robot actuation and transmission mechanisms --~which never evolved for the robot to increase its dynamism substantially~-- nor its force sensing capabilities --~which are derived from Force/Torque sensors of 45 mm diameter installed in the robot \cite{Fumagalli2012}. The iCub3 humanoid robot shown in Fig. \ref{fig:icub_upper_comparison}A is the outcome of a design effort that takes a step in all these directions. The robot represents a concept of humanoid that will be the starting point when conceptualising the next generations of the iCub platform.

\subsubsection*{Mechanics}
The iCub3 humanoid robot is \SI{125}{\cm} tall, and weighs \SI{52}{\kg}. Its mechanical structure is mainly composed by an aluminum alloy. The robot also presents plastic covers that partially cover the electronics. The weight is distributed as follows:  45\% of the weight is {in} the legs, 20\% {in} the arms, and 35\% {in} the torso and head.
Each robot leg is approximately \SI{63}{\cm} long, while the arms are \SI{56}{\cm} long from the shoulder to the fingertips. With the arms along the body, the robot is \SI{43}{\cm} wide. Each foot is composed of two separate rectangular sections, with a total length of about \SI{25}{\cm} and \SI{10}{\cm} wide.

The iCub3 robot possesses in total 54 degrees of freedom including those in the hands and in the eyes, and they are all used in the avatar system. They are distributed as follows: 4 joints in the head controlling the eyelids and the eyes, 3 joints in the neck, 7 joints in each arm, 9 joints in each hand, 3 joints in the torso, 6 joints in each leg.
The iCub3 hands are equipped with tendon driven joints, moved by 9 motors, allowing to control separately the thumb, the index, and the middle finger, while the ring and the pinkie fingers move jointly \cite{schmitz2010design}.

\subsubsection*{Actuation}
The iCub3 is equipped with both DC and brushless three-phase motors.
The DC motors actuate the joints controlling the eyes, the eyelids, the neck, the wrists and the hands. They are equipped with a Harmonic Drive gearbox with 1/100 reduction ratio. 
The torso, the arms and the legs are controlled by three-phase brushless motors, also coupled with {a} 1/100 Harmonic Drive gearbox, with the exception of the hip and ankle roll joints which have a 1/160 gearbox. The motor {characteristics} are as follows. The rated power is \SI{110}{\watt}, with a rated torque of \SI{0.18}{\newton\meter}, while the continuous stall torque is \SI{0.22}{\newton\meter}. 
The hip pitch, knee, and ankle pitch joints are driven by another type of brushless motor, {whose} rated power is \SI{179}{\watt}, {the} rated torque {is} \SI{0.43}{\newton\meter} and {the} continuous stall torque {is} \SI{0.48}{\newton\meter}.

\subsubsection*{Power, Connectivity, Computation, and Electronics}\label{sec:electronics}
The iCub3 robot is powered either by an external supplier or by a custom{-}made battery of \SI{600}{\watt\hour}. The connection to the robot can be established through an Ethernet cable or wirelessly via a standard 5GHz Wi-Fi network.
The robot head is equipped with a 11$^{th}$ generation Intel\textsuperscript{\textregistered} Core i7@\SI{1.8}{\giga\hertz} computer with \SI{16}{\giga\byte} of RAM and running Ubuntu. This central unit represents the interface between the robot and the other laptops in the \emph{robot network}, Fig. \ref{fig:teleoperation_system}.
The iCub3 central unit communicates with a series of boards distributed on the robot body and connected via an Ethernet bus\cite{icub_wiring}. There are two main types of boards connected to the bus, {both 32-bit Arm Cortex micro-controllers}. The first are the \emph{Ethernet Motor Supervisor} (EMS) boards, controlling the three phase motors {with different control strategies. {More details are available online \cite{icub_control_modes}}. They run at \SI{1}{\kilo\hertz} and communicate via CAN protocol with the motor driver board (2FOC), which generates PWM signals at \SI{20}{\kilo\hertz};}. The second are the MC4Plus boards, controlling the DC motors.

\subsubsection*{Sensors}
A particular feature of iCub3 is the vast array of sensors available. iCub3 possesses 8 six-axes force/torque (F/T) sensors \cite{Fumagalli2012} with integrated IMUs. More specifically, there are two different types, F/T-45 and F/T-58, where the number indicates the outer diameter of the sensor. The robot has six F/T-45 sensors. Two of them are mounted at the shoulders, and two on each foot, connecting the two sections of the feet to the ankle assembly. Two F/T-58 are located in the middle of the robot thighs.
iCub3 possesses tactile sensors as an artificial skin \cite{Cannata2008} on the upper arm and the hands.

The head possesses two Basler\textsuperscript{\textregistered} daA3840-30mc cameras capturing images at 30 frames per second, with a 4K resolution. The resolution and the framerate are trimmable to reduce the network load. The images coming from the two sensors are processed by a NVIDIA\textsuperscript{\textregistered} Jetson Xavier NX Module. The cameras are placed within the eyes bulb and can be controlled to a specified vergence, version and tilt angle. Both eyes are equipped with eyelids, controlled jointly by a single DC motor. The robot head also includes a microphone on both ears, and a speaker behind the face cover. Finally, a set of LEDs define the robot face expression.

At the joint level, the iCub3 robot uses a series of encoders. First, an optical encoder mounted on the motor axis {estimates} the motor magnetic flux. Second, the EMS boards exploit an off-axis absolute magnetic encoder mounted on each joint, after the gearbox, to estimate each joint position and velocity.

\subsubsection*{Comparison with the classical iCub platform}

\begin{figure}[pb]
	\centering
	\includegraphics[width=0.8\columnwidth]{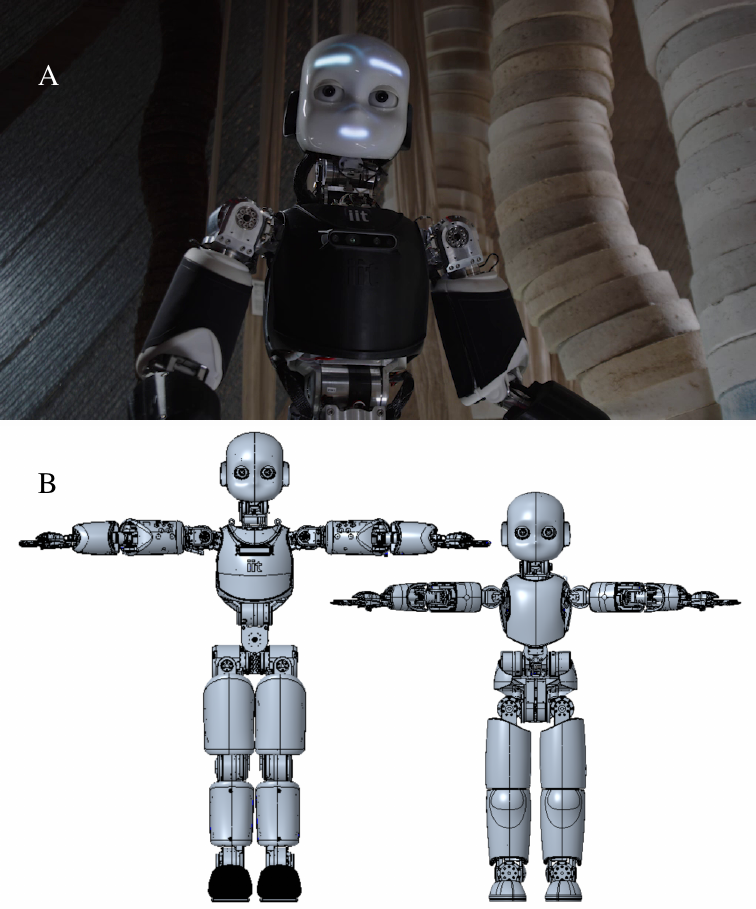}
	\caption{\textbf{iCub3 upperbody and comparison with the previous iCub version}. iCub3 differs from its predecessors, being taller and heavier. iCub3 upperbody (\textbf{A}). Compared to the previous versions, the iCub3 shoulders and torso do not use any tendon-driven mechanism. The iCub3 robot compared to the iCub versions 1.0-2.5 (\textbf{B}).}
	\label{fig:icub_upper_comparison}
\end{figure}

With respect to a classical iCub platform \cite{Nataleeaaq1026}, the iCub3 humanoid robot is {\SI{21}{\cm} taller}, and {weighs} \SI{19}{\kg} more. Fig. \ref{fig:icub_upper_comparison}B shows the different dimensions of the two platforms. The increased weight requires more powerful motors on the legs. 
Moreover, the torso and shoulder joints are serial direct mechanisms, while classical iCub robots have coupled tendon-driven mechanisms. This allows higher range of motion and greater mechanical robustness.

In addition, iCub3 has a higher capacity battery, \SI{10050}{\milli\ampere\hour} versus \SI{9300}{\milli\ampere\hour}, and this is part of the torso assembly instead of being included in a rigidly attached backpack.
{The mechanics of the iCub head and hands have been retained from the classical iCub.}
From the electronics point of view, both platforms share the same 2FOC/EMS/MC4Plus architecture, although iCub3 has higher resolution joint encoders, using 18 bits compared to the 12 of the classical iCub architecture. The PC mounted inside the iCub3 head is more powerful and can also leverage the GPU capabilities of the Jetson Xavier board. 
{The }iCub3 platform has an additional Intel Realsense D435i depth camera placed in the front part of the torso, while the eye cameras have better resolution. In addition, the F/T-58 sensors are only used in the iCub3 robot.

{
\subsubsection*{Robot control}
}

The robot motion is controlled by adopting a layered control architecture~\cite{romualdi2018benchmarking}. Each layer generates references for the layer below by processing inputs from the robot, the environment, and the output of the previous layer.
The inner {the} layer, the shorter the time horizon used to evaluate the output. In addition, lower layers usually employ more complex models to evaluate output, but a shorter time horizon often results in faster computations to obtain these outputs. The mathematical details of this layered architecture are provided in the ``Robot control layered architecture'' section in the Supplementary Materials.

\subsection*{The Communication layer}

Both the robot and the operator system require a cluster of different PCs connected in {two interlinked} local area network{s} (LAN), running multiple applications at once on different operating systems. The communication between the different applications is done through YARP \cite{fitzpatrick2014middle}. 

YARP supports building a robot control system as a collection of programs communicating in a peer-to-peer way, with an extensible family of connection types, like TCP, UDP, or other carriers tailored for the streaming of images.

For real-time operation, network overhead has to be minimized, so YARP is designed to operate on an isolated network or behind a firewall. {However}, the operator and the robot might be {in} two different far places. {To} have the two sub-networks connected, we use OpenVPN\cite{openvpn}.
A simplified diagram of the robot and operator network is depicted in Fig. \ref{fig:teleoperation_system}. {The latency of introduced by the VPN can go from \SI{5}{\milli\second} in a local configuration, to several hundreds of milliseconds in case of bad internet connection.}

\subsection*{The Operator system}

\begin{figure}[tpb]
    \centering
    \includegraphics[width=\columnwidth]{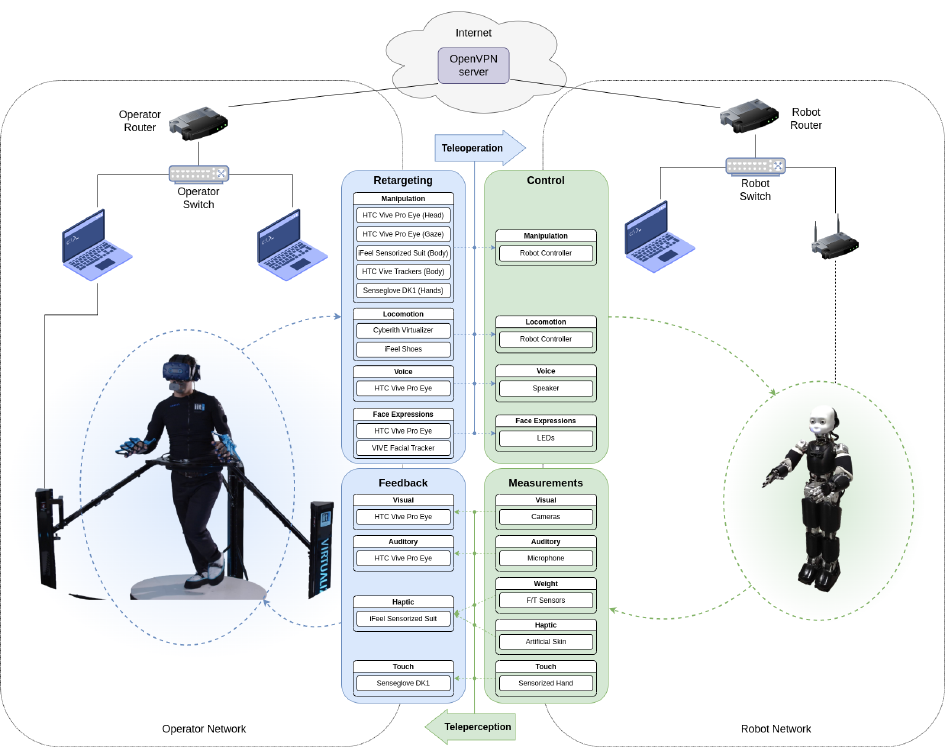}
    \caption{\textbf{The full architecture}. The avatar architecture, comprising the operator, the delayed network, and the avatar. The operator skills are retargeted to the robot through the control architecture, and the operator receives feedback due to the robot sensor measurements.}
    \label{fig:teleoperation_system}
\end{figure}

In the iCub3 avatar system, presented in Fig. \ref{fig:teleoperation_system}, the operator exploits a series of devices. From the HTC VIVE\tm family, we adopt the Pro Eye\tm  headset\cite{htc_pro_eye} with the facial tracker\cite{vive_face_tracker}, and a set of trackers\cite{face_tracker}. The operator also uses the SenseGlove DK1\tm haptic gloves\cite{senseglove} and the Cyberith Virtualizer\tm Elite 2 omnidirectional treadmill\cite{virtualizer}. Finally, the IIT custom-developed iFeel\cite{ifeel} sensorized haptic suit, and shoes complete the set of wearable devices.
The operator devices constitute the \emph{retargeting} and \emph{feedback} interfaces defined in Fig. \ref{fig:teleoperation_system}.

The \emph{retargeting} interfaces contain the set of commands that the operator exploits (on the robot) to achieve a specified task in the remote environment.
In the iCub3 avatar system, we can distinguish the following \emph{retargeting interfaces}: \emph{manipulation}, \emph{locomotion}, \emph{voice} and \emph{face expressions}.

\subsubsection*{Manipulation interfaces}
The \emph{manipulation} interfaces are responsible to process the operator motion to control the robot upper-body. {T}he reference trajectories, fed to the robot controller presented in the ``Robot control'' section, are computed using a multi-modal sensor-fusion algorithm able to combine sensory information from the HTC VIVE\tm headset and trackers, SenseGlove\tm haptic gloves and iFeel nodes.
The headset and the trackers provide position-and-orientation measurements, which are scaled depending on the operator-avatar length ratio and used as a reference for the head and hands motion. 
The iFeel nodes contain an integrated inertial measurement unit (IMU) that measures the gravity vector, orientation, and angular velocity of the associated limbs. Similarly, an IMU is integrated into the haptic gloves, providing orientation and angular velocity of the hands.

The retargeting algorithm is modular and can be scaled depending on the available measurements. It is detailed in the section ``Manipulation interfaces inverse kinematics algorithm'' of the Supplementary Material. Figure \ref{fig:kinematic_retargeting} presents two different sensor configurations used for upper-body motion retargeting on iCub3: iFeel only and iFeel plus trackers. 

\paragraph{iFeel only}\label{par:ifeel_only_retargeting}
{The} headset tracks the motion of the head, while the body motion is controlled exclusively by using the orientation and velocity measurements provided by the nodes, {whose data is acquired at \SI{70}{\hertz}}. This configuration and the corresponding mapping are presented in Figure \ref{fig:kinematic_retargeting}A.

\paragraph{iFeel plus trackers}\label{par:ifeel_plus_trackers}
Since the IMUs estimation can be subject to divergence around the gravity axis and the robot end-effector cartesian position is dependent on the model kinematic chain, VIVE\tm trackers are added to the tracking system. In this configuration{,} gravity information provided by the nodes is used to regulate the internal movements of the robot, while the trackers measure the desired cartesian position for the hands {at \SI{90}{\hertz}}, Figure \ref{fig:kinematic_retargeting}B.

In addition, the operator's gaze and eye openness are tracked using the VIVE headset and facial tracker, allowing to control directly the robot eyelids and gaze \cite{8470602}. The SenseGlove haptic glove completes the set of devices of the \emph{manipulation} interface. It is an exoskeleton-like haptic glove allowing the translation of the motion of each of the operator's fingers into a reference for the robot fingers.

\subsubsection*{Locomotion interfaces}
The \emph{locomotion} interface takes care of detecting the operator's walking intention and commands the robot locomotion. We implement this interface in two different ways: \emph{Virtualizer} and \emph{iFeel Walking}.
\paragraph{Virtualizer}
The Cyberith Virtualizer Elite 2\tm is an omnidirectional treadmill where the operator walks by sliding. The motion is detected through optical sensors located on the device base plate.
The motion direction is estimated via a moving ring attached to the harness secured to the operator's waist. The base plate can also be inclined of a fixed amount to ease the sliding motion, allowing the operator to walk naturally. The walking motion of the operator {generates} a reference walking direction and speed \cite{elobaid2019telexistence}. These references are fed to the planning layer{, }described in the ``Robot control layered architecture'' section of the Supplementary Material, and interpreted as a reference point in Eq.\eqref{eq:unicycle_controller}.

\begin{figure}[tpb]
	\centering
	\includegraphics[width=\columnwidth]{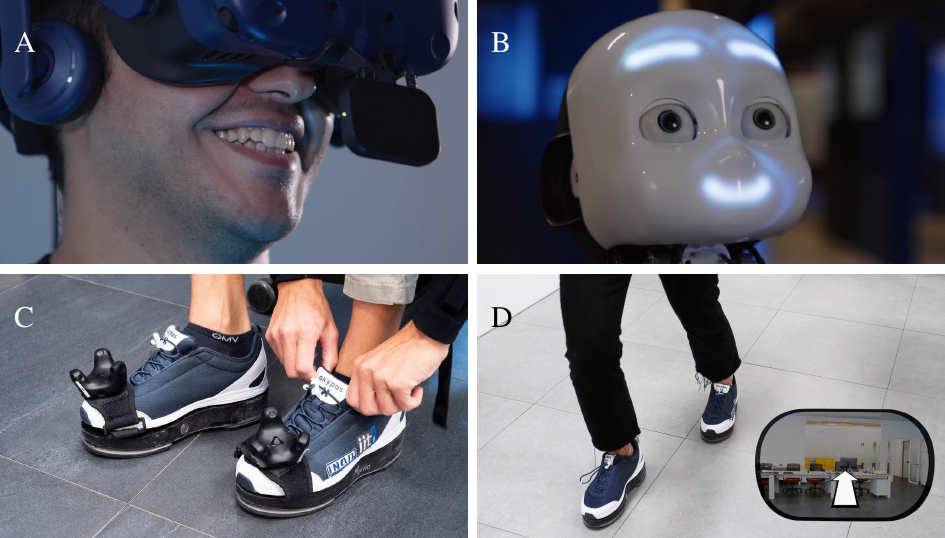}
	\caption{\textbf{Face expressions retargeting and iFeel walking}. Examples of the retargeting interfaces. With the facial tracker (\textbf{A}), the operator can directly control the emotions displayed by the robot (\textbf{B}). The iFeel shoes (\textbf{C}). They measure the force and torque exchanged by the operator with the ground. When paired with a set of trackers, it is also possible to detect their position. An example of the intention mechanism used for the \emph{locomotion} interface (\textbf{D}).}
	\label{fig:operator_Retargeting}
\end{figure}

\paragraph{iFeel Walking}
{T}he Virtualizer platform is {bulky, limiting }its transportability. Moreover, {it is} not possible to command a sideways motion. Hence, we developed the \emph{iFeel walking}. It is composed of two logical components: \emph{intention detection} and \emph{triggering}. The \emph{intention detection} defines the desired locomotion type. More specifically, moving one foot forward or backward enables forward and backward walking, respectively. Contrarily, moving one foot aside enables the lateral walking in the direction of the foot that moved (for example, moving the right foot to the side enables the right sidestepping). Finally, rotating the right (left) foot clockwise (counterclockwise) enables the clockwise (counterclockwise) in-place rotation.
The intention is visualized in the VR headset through a set of arrows, Fig. \ref{fig:operator_Retargeting}D. Then, by stepping in place, the operator triggers the robot's motion in the specified direction. The robot's desired walking speed is modulated by the stepping frequency. Each intention is mapped to the corresponding control input on the modified unicycle dynamics of Eq.\eqref{eq:unicycleDynamicsModified}.

The iFeel walking system requires measuring the relative position of each operator's foot with respect to the waist, and the normal force exerted in each foot to detect the stepping. The first quantity is measured via a set of VIVE\tm trackers on the operator's feet and waist. The second quantity, instead, is measured thanks to the iFeel shoes, shown Fig. \ref{fig:operator_Retargeting}C. The iFeel shoes estimate the interaction forces exchanged by the operator with the ground by means of two F/T-45 sensors installed on the soles. 

Compared to Virtualizer solution, the \emph{iFeel walking} solution does not constrain the operator at a fixed point. This might disorient some novice operators, as the immersivity can affect their sense of equilibrium. Moreover, the operator could step away from the tracked area. In order to avoid this issue, a message is printed on the headset to suggest the operator to move back to the original location.

\subsubsection*{Voice and Face Expressions interfaces}
The \emph{voice} and \emph{face expressions} interfaces exploit the HTC VIVE\tm headset microphone and the attached VIVE\tm facial tracker. The former allows the operator to verbally interact through the robot. The latter is fundamental for the non-verbal interaction. 
{Thanks to the headset facial tracker, the operator's face expressions }are replayed by the robot LEDs, Fig. \ref{fig:operator_Retargeting}(A and B).

\subsubsection*{The feedback interfaces}
The \emph{feedback} interfaces report the robot sensors measurement to the operator. In the iCub3 teleoperation system we have the following \emph{feedback} interfaces: \emph{visual}, \emph{auditory}, \emph{haptic}, \emph{touch}.
The headset is fundamental for the \emph{visual} and \emph{auditory feedback}. The images captured by the robot cameras are displayed inside the headset.
At the same time, the audio captured by the robot microphones is directly played on the headset's headphones.
The SenseGlove\tm haptic gloves provide \emph{touch} feedback by means of vibration motors in each fingertip{,} on the back of the hand{,} and through a set of brakes {that} produce up to \SI{20}{\newton} of passive force per finger.
The iFeel haptic nodes are fundamental for the body \emph{haptic feedback}. They {are} used in two different ways: \emph{touch feedback} and \emph{weight feedback}.

\paragraph{Touch feedback}
The iFeel haptic nodes reproduce a touch occurring on the robot arm. {The} sensorized skin mounted on the robot arms {detects} a contact that is reproduced on the operator's arms through a vibration. 

\paragraph{Weight feedback}
We use the iFeel haptic nodes also to retarget the effort endured by the robot arms. In particular, the haptic nodes modulate the vibration differently according to the amount of vertical force exerted on each {robot} arm.

\subsection*{Logging systems}

{We} implemented two logging systems for two different purposes: \emph{online} monitoring and \emph{offline} processing.
The online logging mechanism exploits the \texttt{openmct}\cite{openmct} framework to display the data measured from the robot. 
It connects through YARP reading the robot data streams, making it available from a normal browser{, also from personal mobile devices. }
The code is available online\cite{yarp_openmct}. Figure \ref{fig:Openmct}(A) shows an example visualization of the GUI, with a live plot of the battery status of charge, and the communication delay to the robot PC. 

The data streamed by the robot, together with some additional data coming from the walking controller, is also saved periodically in \texttt{.mat} files for \emph{offline} analysis: the code is open-source and available online\cite{robometry}. We implemented the so-called \texttt{robot-log-visualizer}\cite{robot_log_visualizer} to quickly visualize and plot such data. Inspired by IHMC's SCS \cite{pratt2009yobotics}, \texttt{robot-log-visualizer} allows visualizing the data by simply clicking on the data of interest in the left panel, as shown in Fig. \ref{fig:Openmct}(B). On the right panel, we have a 3D representation of the robot. If available, it is also possible to display a synchronized camera stream.

\section*{Supplementary Materials and Methods}
The Supplementary Materials contain the mathematical derivations of the ``Robot control'' and the ``Manipulation interfaces`` in the sections named ``Robot control layered architecture'' and ``Manipulation interfaces inverse kinematics algorithm'', respectively. It also contains Figs. \ref{fig:semifinals_puzzle} to \ref{fig:Openmct}.

\bibliographystyle{IEEEtran}

\section*{Acknowledgments}
This system paper was made possible by the effort of many people that contributed at different stages of the project. In particular, we would like to acknowledge the help of Mattia Salvi, Davide Tomè, Prashanth Ramadoss, Yeshasvi Tirupachuri, Diego Ferigo, Raffaello Camoriano, Guglielmo Cervettini in making this paper a reality.

\paragraph{Funding:} The paper was supported by the Italian National Institute for Insurance against Accidents at Work (INAIL) ergoCub Project.

\paragraph{Author contributions:} S. Dafarra lead the integration and validation activities. U. Pattacini lead the activities for the preparation and enhancements of the iCub3 robot. G. Romualdi developed the robot locomotion algorithms. L. Rapetti developed the operator retargeting algorithms. R. Grieco developed the weight retargeting application and contributed to the iFeel walking algorithm. K. Darvish contributed to the telexistence system algorithms/integration and developed the anthropomorphic telemanipulation framework. G. Milani contributed to the development of the iFeel hardware. E. Valli supervised the development of the iFeel hardware. I. Sorrentino improved the robot's low-level control and estimation of external forces. P. M. Viceconte developed the texture classification algorithm. A. Scalzo developed the iCub3 low-level control. S. Traversaro contributed to the architecture software stack and developed the estimation of the iCub3 joint torques. C. Sartore developed the iFeel walking algorithm. M. Elobaid developed the first teleoperation architecture. N. Guedelha developed the online logging visualization tool. C. Herron developed the first version of the iFeel walking. A. Leonessa supervised the development of the first version of the iFeel walking. F. Draicchio contributed to design iteration of the iFeel wearables. G. Metta conceived the original idea of the iCub3 robot. M. Maggiali lead the development of the iCub3 robot. D. Pucci supervised all the activities.

\paragraph{Competing interests:} The authors declare that they have no competing interests.
\paragraph{Data and materials availability:} All data needed to support the conclusions of this manuscript are included in the main text {and Supplementary Materials. The scripts to generate Figures \ref{fig:iCub3_wmf}(C-D) and \ref{fig:semifinals}{(D-H)} are provided as separate zip file.}
\newpage
\section*{{Figures and Tables}}

\begin{table*}[bp]
\newcolumntype{Y}{>{\centering\arraybackslash}X}
\centering
\caption{\textbf{Summary of the set of validations we used for the iCub3 avatar system}. For each validation, we define a set of requirements that meet specific objectives{.}
More specifically, a validation might require the avatar to be in a remote location with respect to the operator (\emph{Remote}), or at a close distance (\emph{Local}), typically in the same building. Another requirement is represented by the level of expertise of the operator. We define an \emph{Expert} operator someone that has deep knowledge of the  avatar system, while a \emph{Naive} operator has to be trained before the beginning of the validation -- the training time is fixed at about thirty minutes. We also categorize the validations according to the skills required on the avatar, in terms of locomotion, interaction, and manipulation. Given the requirements and objective, we show which avatar system might be implemented. Each \emph{System Setting and Algorithm} is detailed in a specific paragraph in the Methods section.}
\label{tab:validation_requirements}
\resizebox{\textwidth}{!}{\begin{tabularx}{900pt}{c *{12}{Y} *{4}{c}}
\toprule
\multirow{3}{*}[-45pt]{Validation} & \multicolumn{12}{c}{Requirements} &  \multirow{3}{*}[-45pt]{Objectives} & \multicolumn{3}{c}{System Settings and Algorithms}\\
\cmidrule(lr){2-13} \cmidrule(l){15-17}
 & \multicolumn{2}{c}{Location} & \multicolumn{2}{c}{Operator} & \multicolumn{3}{c}{\makecell{Locomotion\\Capabilities}} & \multicolumn{3}{c}{\makecell{Interaction with\\Recipient}} & \multicolumn{2}{c}{Manipulation} & & \multirow{2}{*}[-30pt]{Tracking System} & \multirow{2}{*}[-30pt]{\makecell{Body Haptic\\Feedback}} & \multirow{2}{*}[-30pt]{\makecell{Locomotion\\Retargeting}} \\
\cmidrule(lr){2-3}\cmidrule(lr){4-5}\cmidrule(lr){6-8}\cmidrule(lr){9-11}\cmidrule(lr){12-13}
 & \rotatebox[origin=c]{90}{Local} & \rotatebox[origin=c]{90}{Remote} & \rotatebox[origin=c]{90}{Expert} & \rotatebox[origin=c]{90}{Naive} & \rotatebox[origin=c]{90}{Short Distance} & \rotatebox[origin=c]{90}{~~Long Distance~~~} & \rotatebox[origin=c]{90}{Side Motions} & \rotatebox[origin=c]{90}{Verbal} & \rotatebox[origin=c]{90}{~~Non-Verbal~~~} & \rotatebox[origin=c]{90}{Physical} & \rotatebox[origin=c]{90}{Coarse} & \rotatebox[origin=c]{90}{Precise} & & & & \\
\midrule
Italian Pavillion &  & \checkmark & \checkmark & & & \checkmark & & \checkmark & \checkmark & \checkmark & \checkmark & & \makecell{Physical and\\non-verbal interaction} & iFeel Only & Touch Feedback & Virtualizer\\[15pt]
We Make Future & & \checkmark & \checkmark & & \checkmark & & & \checkmark & \checkmark & \checkmark & & \checkmark & \makecell{Physical collaboration and\\public engagement} & iFeel + Trackers & Weight Feedback & Virtualizer\\[15pt]
XPrize Semifinals & \checkmark & & & \checkmark & \checkmark & & & \checkmark & \checkmark &  &  & \checkmark & \makecell{Fine manipulation and\\shared situational awareness} & iFeel + Trackers & Weight Feedback & Virtualizer\\[15pt]
XPrize Finals & \checkmark &  &  & \checkmark &  &\checkmark & \checkmark & \checkmark &  &  &  & \checkmark & \makecell{Mission oriented\\loco-manipulation} & iFeel + Trackers & Weight Feedback & iFeel Walking \\
\bottomrule
\end{tabularx}}
\end{table*}

\begin{table*}
\centering
\caption{\textbf{List of the ANA Avatar XPrize Finals tasks}. Each task had to be completed sequentially. Failing to complete one task caused the end of the scored trial.}
\label{tab:finals_tasks}
\begin{tabularx}{\textwidth}{c}
\toprule
Tasks\\
\midrule
\makecell{The Avatar walks about 5 meters to a designated spot, allowing the Operator\\to communicate with the Mission Commander, who explains the mission.}\\[15pt]
The Avatar walks about 5 meters and activates a switch that opens the station door.\\[15pt]
\makecell{The Avatar walks about 30 meters to the next task, where it has to identify\\one heavy canister according to its weight and place it in a designated spot.}\\[15pt]
\makecell{The Avatar walks about 10 meters between obstacles, up to a table with a drill.\\ The Avatar activates the drill and unscrews a pin holding an opening with a small curtain.}\\[15pt]
The Avatar reaches through the curtain to identify a rough textured rock and retrieve it.\\
\bottomrule
\end{tabularx}
\end{table*}

\vfill

\newpage
\appendix
\setcounter{page}{1}
\setcounter{figure}{0}
\setcounter{equation}{0}
\title{iCub3 Avatar System: \\
Enabling Remote Fully-Immersive\\
Embodiment of Humanoid Robots\\[15pt]
Supplementary Material}
\maketitle
\newpage
\renewcommand{\theequation}{S\arabic{equation}}
\section*{Robot control layered architecture}
The layered control architecture mentioned in the ``Robot control'' section of the ``Methods'' is composed of three layers. From top to bottom, the layers are here called: \emph{trajectory optimization}, \emph{simplified model control}, and \emph{whole-body quadratic programming (QP) control} -- Figure~\ref{fig:three-layer}.

\paragraph{Trajectory optimization} The \emph{trajectory optimization} {layer} aims to compute a sequence of contacts' location and timings. This layer often takes advantage of optimization techniques to consider the feasibility of the contact location. 
The simpler the model, the simpler the problem. 
For instance, flat terrain allows one to model the robot as a simple unicycle~\cite{flavigne2010reactive} which enables fast solutions to the optimization problem for the walking pattern generation~\cite{8594277}.\\

The unicycle model is described by the following equations \cite{pucci2013nonlinear}:
\begin{IEEEeqnarray}{RCL}
\IEEEyesnumber \phantomsection \label{eq:unicycleDynamics}
\dot x_u & = & v_u R_2(\theta_u) e_1, \IEEEyessubnumber \label{eq:honolonimicCon} \\
\dot \theta_u & = & \omega_u,       \IEEEyessubnumber
\end{IEEEeqnarray}
with $v_u \in \mathbb{R}$ and $\omega_u \in \mathbb{R}$ the unicycle's rolling and rotational velocity, respectively. $x_u \in \mathbb{R}^2$ is the unicycle position in the inertial frame $\mathcal{I}$, while $\theta_u \in \mathbb{R}$ represents the angle around the $z-$axis of $\mathcal{I}$ which aligns the inertial reference frame with a unicycle fixed frame.
$R_2(\theta) \in SO(2)$ is the rotation matrix of an angle $\theta \in \mathbb{R}$ in a 2D plane, while $e_1 = [1, 0]^\top$.

A possible control objective for this kind of model is to asymptotically stabilize a point $F$ rigidly attached on the unicycle about a desired point $F^*$ whose position is $x_F^*$. {D}efine the error $\tilde{x}$ as 
\begin{equation}
\tilde{x}_u  := x_F - x_F^* =  x_u + R_2(\theta_u)d_u - x_F^*, \label{eq:xtilde}
\end{equation}
where $d_u \in \mathbb{R}^2$ is the position of $F$ in the unicycle frame. The following control law makes the origin of the error dynamics an asymptotically stable equilibrium \cite{8594277}:
\begin{equation}\label{eq:unicycle_controller}
\begin{bmatrix}v_u \\ \omega_u \end{bmatrix} = \begin{bmatrix}R_2(\theta_u) e_1 &  R_2(\theta_u + \pi/2) d_u\end{bmatrix}^{-1}(\dot{x}^*_F - K_u\tilde{x}_u), 
\end{equation}  
with $K_u$ a positive definite matrix.

In our context, the humanoid robot feet are represented by the unicycle wheels, and the desired footsteps can be obtained by sampling the unicycle trajectories given a desired trajectory for the point $F$. 
The generation of footsteps via the unicycle model allows for planning walking motions using a simple two-dimensional quantity. On the other hand, the unicycle model of Eq.\eqref{eq:unicycleDynamics} does not allow motion along the wheel axis. In other words, the robot would not be able to perform lateral steps. To circumvent this limitation, we modify Eq. \eqref{eq:honolonimicCon} as follows:
\begin{equation}\label{eq:unicycleDynamicsModified}
\dot x_u = R_2(\theta_u) \begin{bmatrix} v_u \\ l_u	\end{bmatrix},
\end{equation}
where $l_u \in \mathbb{R}$ is an additional control input enabling the unicycle side motions. Consequently, rather than employing the control law presented in Eq. \eqref{eq:unicycle_controller}, we directly define the three control inputs $v_u$, $\omega_u$, and $l_u$ according to the desired robot motion. 

Once the footsteps are planned, the desired feet trajectory is obtained by cubic spline interpolation. Assuming a constant height of the center of mass while walking and a constant angular momentum, we plan the center-of-mass (CoM) and zero moment point (ZMP)~\cite{vukobratovic2004zero} trajectory through the \emph{algebric Divergent Component of Motion generator}~\cite{Englsberger2014,romualdi2018benchmarking}.

\paragraph{Simplified model control} The output of the trajectory optimization layer feeds the \emph{simplified model control} {layer} which is responsible for finding feasible robot center-of-mass (CoM) trajectories. 
{Given a} desired CoM $x_\text{CoM}^\text{ref}$ and ZMP $x_\text{ZMP}^\text{ref}$ position to stabilize, we develop a control law that approximates the robot {dynamics} via the Linear Inverted Pendulum Model (LIPM)~\cite{Kajita2001} following \cite{Choi2007}:
\begin{equation}
\label{eq:ZMP_controller}
\dot{x}^*_{\text{CoM}} = \dot{x}^\text{ref}_{\text{CoM}} - K_\text{ZMP}(x_\text{ZMP}^\text{ref} - x_\text{ZMP}) + K_\text{CoM} (x^\text{ref}_{\text{CoM}} - x_{\text{CoM}}),
\end{equation}
where $K_\text{CoM} -\zeta I_2$ {and $ K_\text{ZMP}$ are positive} definite, {while} $ K_\text{ZMP} - \zeta I_2$ is negative definite, with $\zeta \in \mathbb{R}_{>0}$. 

\paragraph{Whole-Body control} The \emph{whole-body control} layer generates joint {position }references {for} the robot.
The proposed controller evaluates the generalized robot velocity $\nu \in \mathbb{R}^{n_s + n_b}$ where $n_s$ is the number of the robot joints and $n_b = 6$ represents the degrees of freedom associated with the floating base.
Recalling that the velocity of a link $L$ depends linearly on $\nu$ employing the Jacobian $J_L$, we define a set of tasks $\Psi_{L_\text{SE(3)}}$ of the form 
\begin{equation}
\label{eq:ik_se3_task}
\Psi_{L_\text{SE(3)}} = \mathrm{v}_L^* - J_L \nu,
\end{equation}
where $\mathrm{v}_L^*$ is the desired velocity chosen to guarantee the tracking of the reference link {pose}~\cite{romualdi2018benchmarking}.
{The SO(3) task, $\Psi_{L_\text{SO(3)}}$, ensures the convergence of a frame orientation to a desired orientation by selecting the appropriate rows from Eq.\eqref{eq:ik_se3_task}.}
\par
While teleoperating, we always require the center of mass $x_\text{CoM}$ to remain in a given position{:}
\begin{equation}
\label{eq:ik_com_task}
\Psi_{\text{CoM}} = \dot{x}_\text{CoM}^* - J_\text{CoM} \nu,
\end{equation}
where $\dot{x}_\text{CoM}^*$ is the desired CoM velocity chosen to guarantee the convergence of the CoM to a given trajectory, while $J_\text{CoM}$ is the linear component of the Centroidal Momentum matrix scaled by the total mass of the robot~\cite{orin08}.
\par
To consider the desired robot joint positions provided by the retargeting system, we introduce a regularization task for the joint variables. The task is achieved by asking for desired joint velocities that depend on the error between the desired and measured joint values, such as:
\begin{equation}
\label{eq:ik_s_task}
\Psi_s = \dot{s}^* - \begin{bmatrix}
0_{n_s\times6} & I_n 
\end{bmatrix} \nu,
\end{equation}
where $n_s$ is the robot actuated degrees of freedom and $\dot{s}^*$ {guarantees} the tracking of the joint reference obtained by the algorithms presented in the ``Manipulation interfaces'' section.

The tracking of the left and right foot poses is considered as high-priority $\SE(3)$ tasks, Eq. \eqref{eq:ik_se3_task}, {that} are denoted as $\Psi_{L_{\SE(3)}}$ and $\Psi_{R_{\SE(3)}}${, respectively}. We take into consideration the CoM tracking as a high-priority task, Eq. \eqref{eq:ik_com_task}.
The torso orientation is considered as a low-priority task $\SO(3)$ task and we denote it with $\Psi_{T_{\SO(3)}}$. The retargeting joint positions tracking is considered as a low-priority regularization task, Eq. \eqref{eq:ik_s_task}, and denoted as $\Psi_{s_{\text{ret}}}$. Furthermore, the joint postural condition, Eq. \eqref{eq:ik_s_task}, is also enforced while walking as a low-priority task,  $\Psi_{s_{\text{reg}}}$.
The above hierarchical control objectives is framed into a whole-body optimization problem:

\begin{IEEEeqnarray}{CL}
\phantomsection \label{eq:ik_optimization} \IEEEyesnumber \IEEEyessubnumber*
\;\minimize\limits_{\nu} \; & \Psi_{T_{\SO(3)}}^\top \Lambda_T \Psi_{T_{\SO(3)}} + \Psi_{s_{\text{reg}}}^\top \Lambda_{s_{\text{reg}}} \Psi_{s_{\text{reg}}}  + \Psi_{s_{\text{ret}}}^\top \Lambda_{s_{\text{ret}}} \Psi_{s_{\text{ret}}}   \label{eq:ik_optimization_cost} \\
\st & \Psi_{L_{\SE(3)}} = 0  \label{eq:ik_optimization_costraint_lf} \\
&  \Psi_{R_{\SE(3)}} = 0 \label{eq:ik_optimization_costraint_rf} \\
& \Psi_{\text{CoM}} = 0 \label{eq:ik_optimization_costraint_com} 
\end{IEEEeqnarray}

The performance of Eq. \eqref{eq:ik_optimization} depends on the choice of the weights $\Lambda_{s_{\text{reg}}}$ and $\Lambda_{s_{\text{ret}}}$. In particular, we observe that the weights achieving good embodiment during standing and walking are not the same. For this reason, we implement a gain-scheduling technique depending on whether the robot is walking or standing. The transition between the two sets of weights is smooth with a minimum acceleration trajectory.

Since the decision variable is the robot velocity $\nu$ and the tasks depend linearly on $\nu$, we transcribe the optimization problem into a quadratic programming (QP) problem, and we solve it via off-the-shelf solvers. The transcription is achieved {through} the Inverse Kinematics implemented in the \texttt{bipedal-locomotion-framework} library\cite{blf}. {The} QP problem is solved by means of \texttt{osqp-eigen}\cite{osqp_eigen} a \texttt{C++} wrapper for OSQP~\cite{Stellato2018}.

\section*{Manipulation interface inverse kinematics algorithm}
The manipulation interfaces use the wearable sensor's measurement altogether as inputs for a constrained inverse kinematics algorithm \cite{rapetti2020model}, mapping the motion into the robot model following the geometric retargeting approach presented in \cite{darvish2019whole}. Depending on the measurement type, the desired model link velocities are defined as follows. For a position $\hat{p} \in \mathbb{R}^3$ and/or a linear velocity measurement $\hat{v} \in \mathbb{R}^3$, the corrected link linear velocity is computed as
\begin{equation}
v^{*} = \hat{v} + K_p  \left ( \hat{p} - p \right ).
\end{equation}
Instead, for an orientation $\hat{R} \in \SO(3)$ and/or an angular velocity measurement $\hat{\omega} \in \mathbb{R}^3$, the corrected link angular velocity is computed as
\begin{equation}
\omega^{*} = \hat{\omega} + K_R \; \text{sk}(R^T  \hat{R})^{\vee},
\end{equation}
where $\text{sk}(.)^{\vee}:\SO(3) \to \mathbb{R}^3 $ is the operator that extracts the skew-symmetric part of the matrix and applies the inverse of the skew operator.
Finally, for a gravity versor measurement $\hat{u}_g \in \mathbb{R}^3, ||\hat{u}_g||=1$, the corrected link angular velocity is computed as
\begin{equation}
\omega^{*} = R (\hat{u}_g \times u_g).
\end{equation}

The corrected link velocities compensate for the measurement error and are achieved by means of the same task formulation presented in Eq.\eqref{eq:ik_se3_task}. In particular, all the corrected link velocities and the respective Jacobians are vertically concatenated into $\mathrm{v}^{*}$ and $J$ respectively,
\begin{equation}
\Psi = \mathrm{v}^{*} - J  \nu.
\end{equation}
Then, we formulate a constrained inverse differential kinematics optimization:
\begin{IEEEeqnarray}{RCL}
\nu^{*} = & \; \underset{\nu}{\text{minimize}} \; &
\Psi^\top \Lambda \Psi \\
& \text{subject to} &   A \nu \leq b,
\end{IEEEeqnarray}
with $\Lambda$ being a weight matrix. $A$ and $b$ are defined to ensure both velocity and configuration constraints using the joint limit avoidance approach described in \cite{rapetti2020model}. The output system velocity $\nu^{*}$ is integrated to obtain the model configuration $q$. {The joint limit avoidance mechanism and the integration passage naturally filter disturbances coming from the input sensors. In fact, the former limits the maximum model joint velocities, while the latter naturally filters high-frequency noise. At the same time, this behavior introduces some lag in the motion retargeting, hence it is necessary to tune $\lambda$ to trade-off between reactivity and data filtering. }

\newpage
\section*{Supplementary figures}
\renewcommand{\thefigure}{S\arabic{figure}}

\begin{figure}[bp]
    \centering
    \includegraphics[width=0.9\columnwidth]{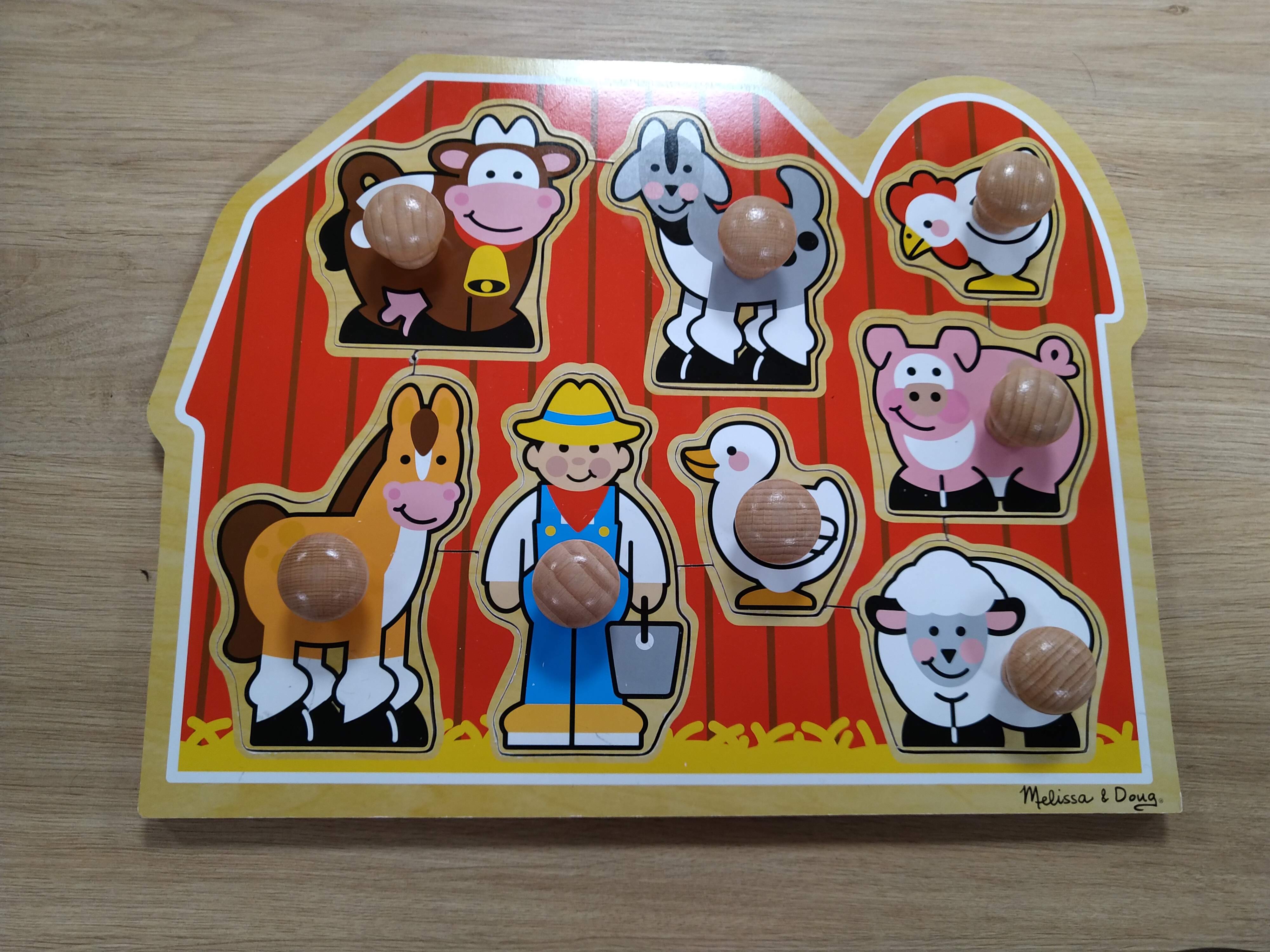}
    \caption{\textbf{The puzzle used during the XPrize semifinals}. Each puzzle piece has a knob that helps their grasping.}
    \label{fig:semifinals_puzzle}
\end{figure}

\begin{figure}[tbp]
	\centering
	\includegraphics[width=0.9\columnwidth]{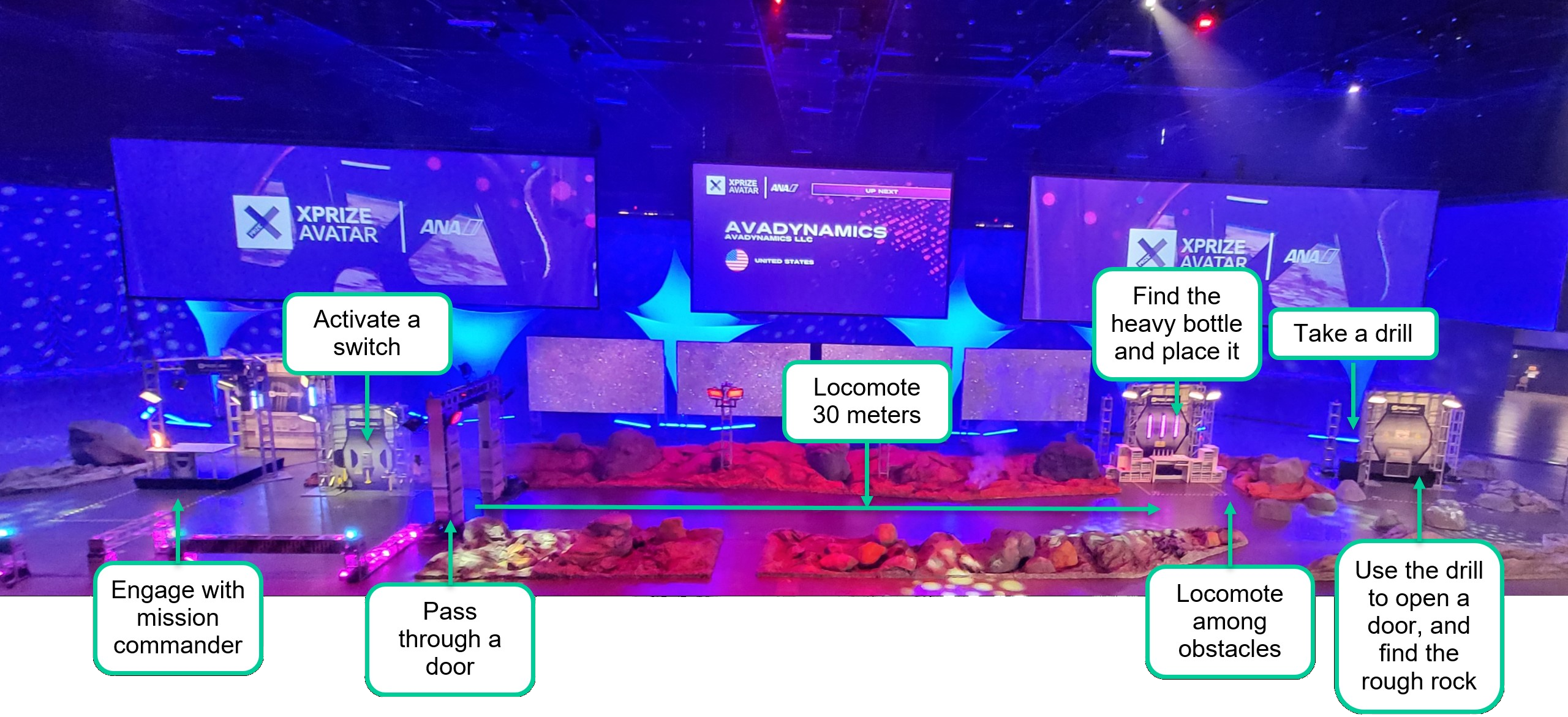}
	\caption{\textbf{A representation of the ANA Avatar XPrize finals course}. It is themed on the exploration of another planet. The labels indicate how the tasks listed in Table \ref{tab:finals_tasks} were executed on the course. }
	\label{fig:finals_stage}
\end{figure}

\begin{figure}[tbp]
    \centering
    \subfloat{\begin{overpic}[width=0.5\columnwidth]{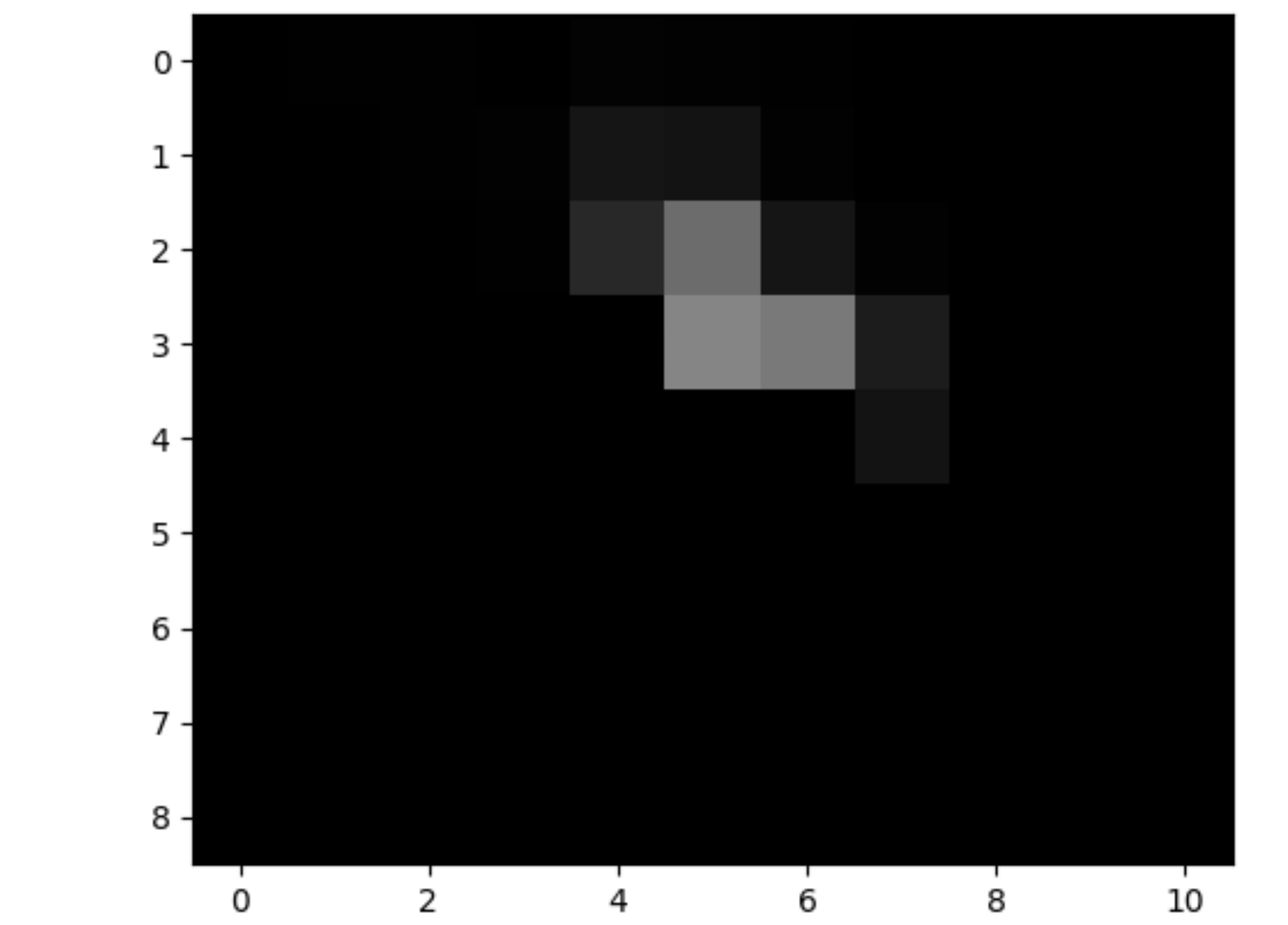}
              \put(2,66){\textcolor{black}{\large{A}}}
              \end{overpic}}
    \subfloat{\begin{overpic}[width=0.5\columnwidth]{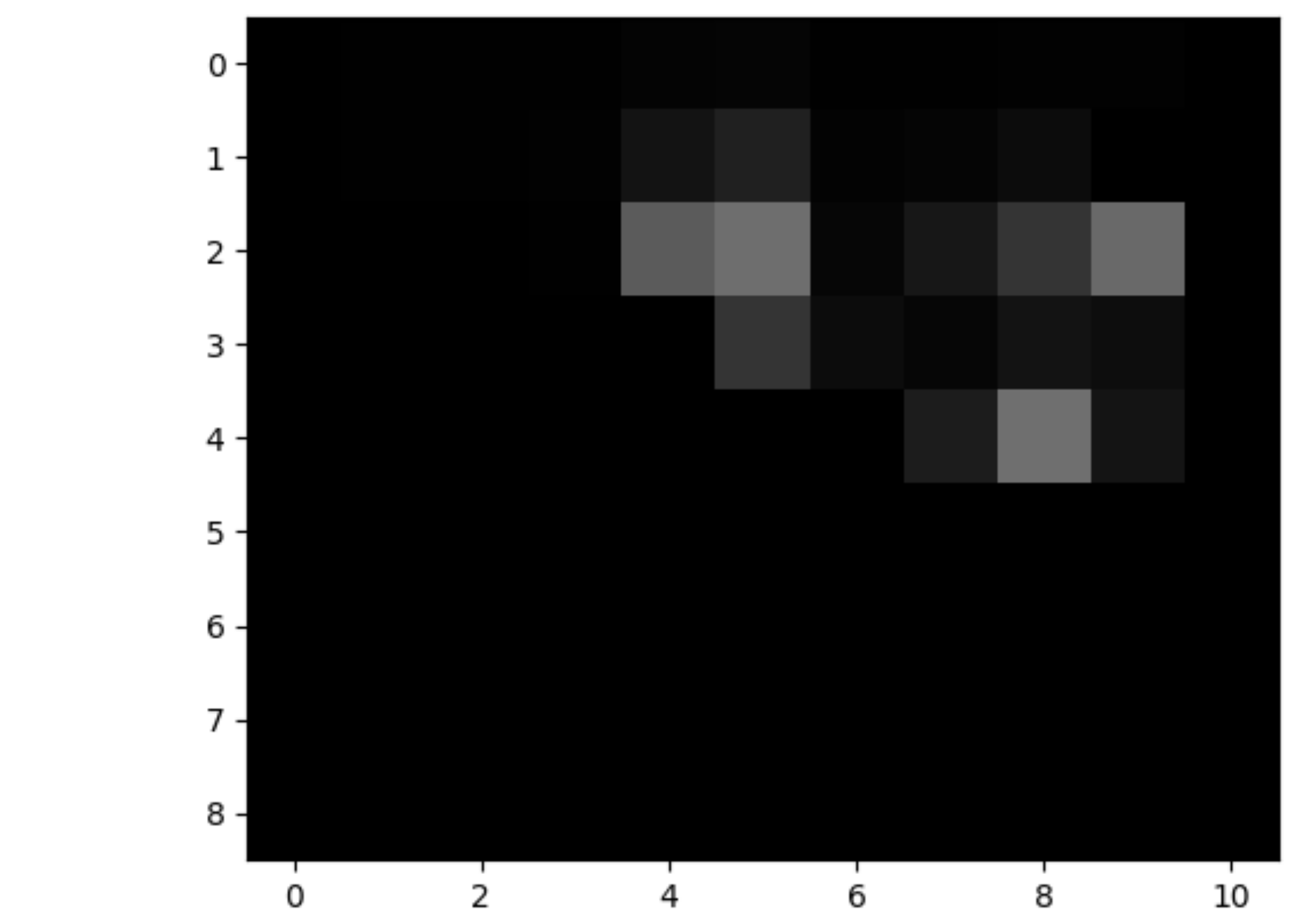}
              \put(6,66){\textcolor{black}{\large{B}}}
              \end{overpic}}\\[15.0pt]
    \subfloat{\begin{overpic}[width=1.0\columnwidth]{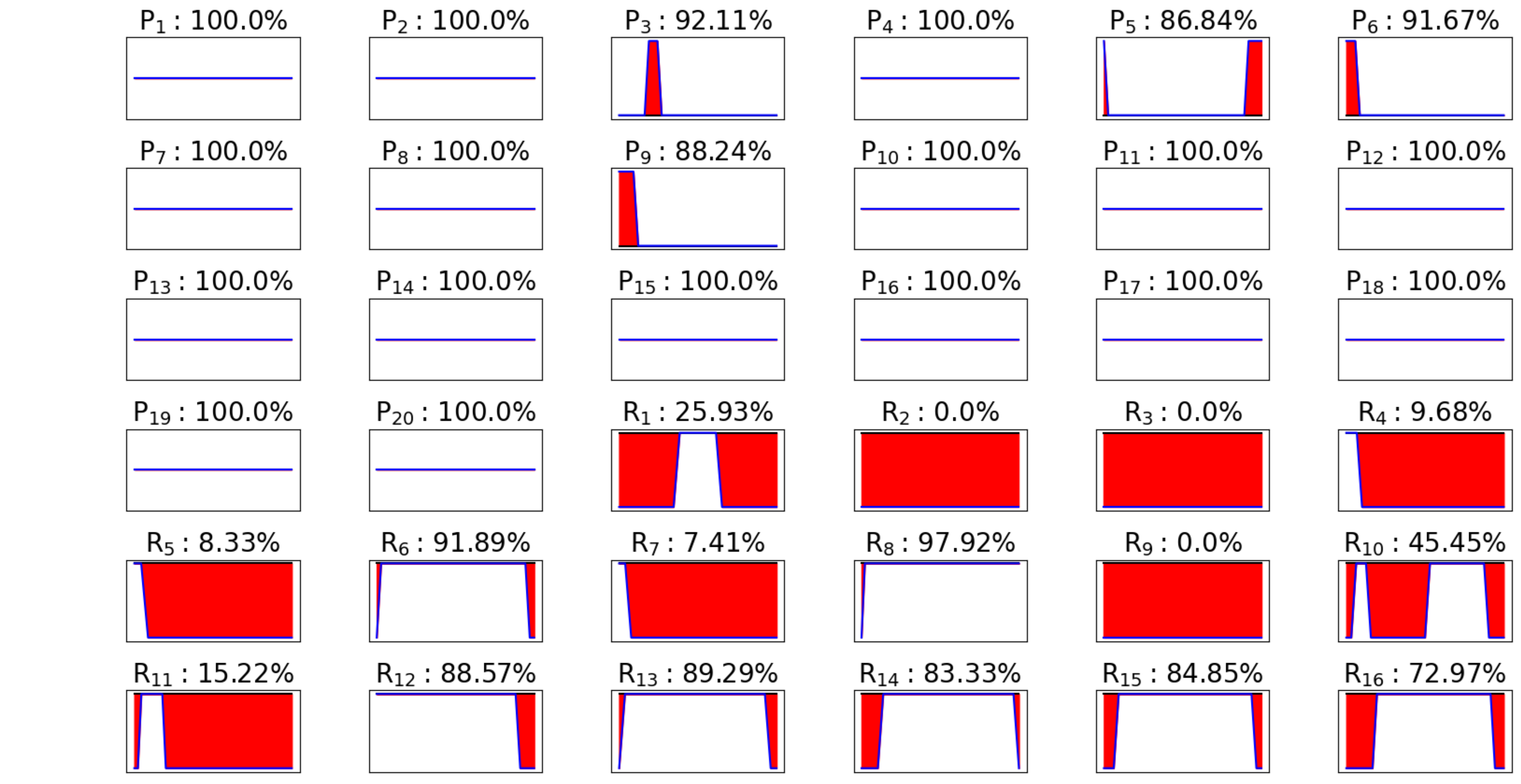}
              \put(1,49){\textcolor{black}{\large{C}}}
              \end{overpic}}
    \caption{\textbf{Performance of the classifier for the XPrize finals texture task}. The images represent the input and the performance of the neural network used to identify the texture of rocks. Grayscale images extracted from the sensorized robot palm in contact with a plain (\textbf{A}) and a rough (\textbf{B}) rock. The higher the pressure measured by each tactile sensor, the whiter the correspondent pixel. The active tactile sensors are more sparse during contact with a rough rock. The rock classifier's performances on the 36 contacts of the test set (\textbf{C}). Each plot corresponds to a separate contact, either plain $\text{P}_\text{i}$ or rough $\text{R}_\text{i}$, and shows a comparison of the ground truth and the predicted class for the entire duration of the contact. When the prediction and the ground truth do not coincide, the gap is filled in red to highlight the error. Although the misclassification rate increases for rough contacts, the overall accuracy on the test set, more precisely the average of the per-contact accuracies labeling each plot, reaches 78\%.}
    \label{fig:texture_task_details}
\end{figure}

\begin{figure}[t]
    \centering
    \includegraphics{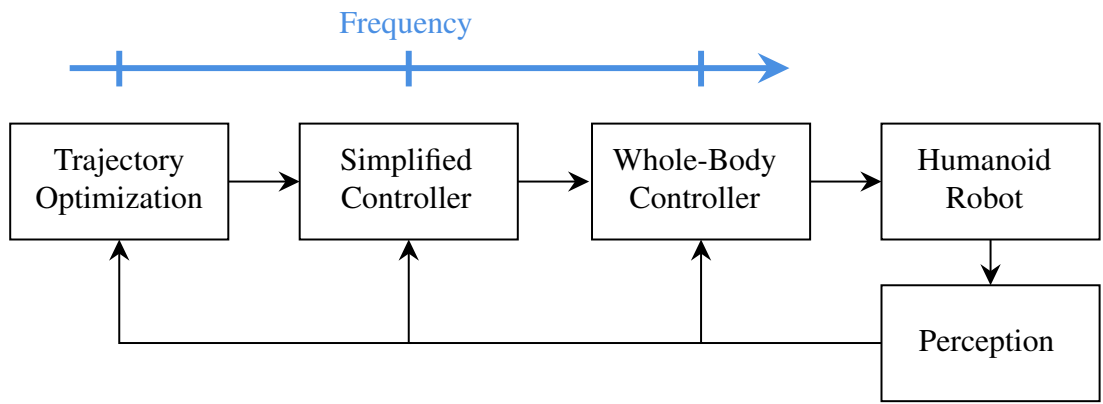}
    \caption{\textbf{The three layer controller architecture}. The inner the layer, the higher the frequency. Each layer gathers the outcome of the outer layer, the information from the robot through the perception block, and generates the references for the inner layer.}
    \label{fig:three-layer}
\end{figure}

\begin{figure}[tpb]
    \centering
    \subfloat{\begin{overpic}[width=0.9\columnwidth]{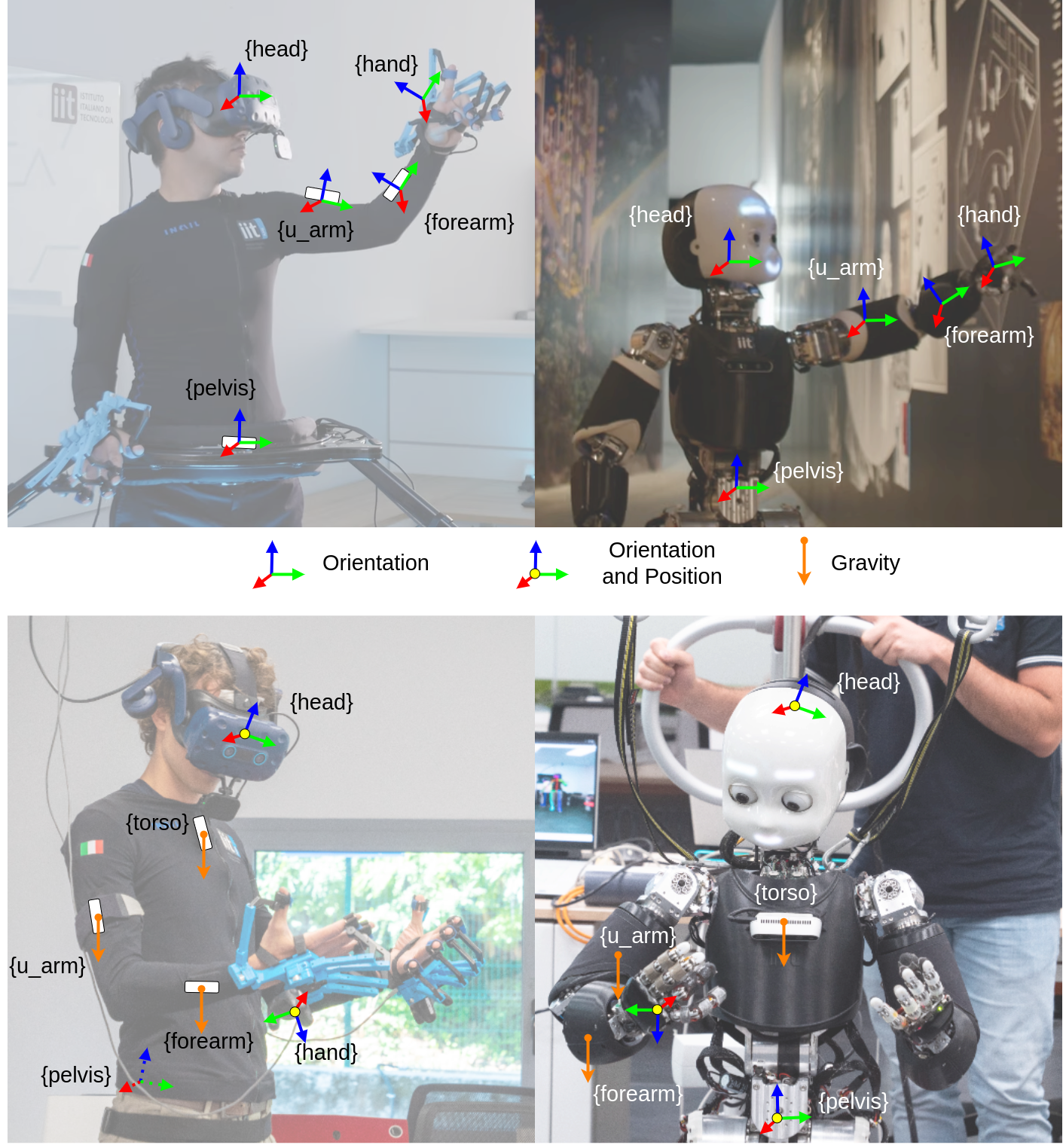}
              \put(3,95){\textcolor{black}{\large{A}}}
              \put(3,40){\textcolor{black}{\large{B}}}
              \end{overpic}}
    \caption{\textbf{Manipulation interfaces}. The operator movements are tracked by the robot through the estimation of a series of ``targets'', as explained in the ``Manipulation interfaces'' section. Upper body motion retargeting using in the ``iFeel only`` configuration  (\textbf{A}).  Upper body motion retargeting using iFeel nodes, headset, and trackers, corresponding to the ``iFeel plus trackers'' configuration (\textbf{B}). In both cases, the same sensor configuration applies to both arms, hence we show the configuration of one arm only.}
    \label{fig:kinematic_retargeting}
\end{figure}

\begin{figure}[tpb]
    \centering
    \subfloat{\begin{overpic}[width=\columnwidth]{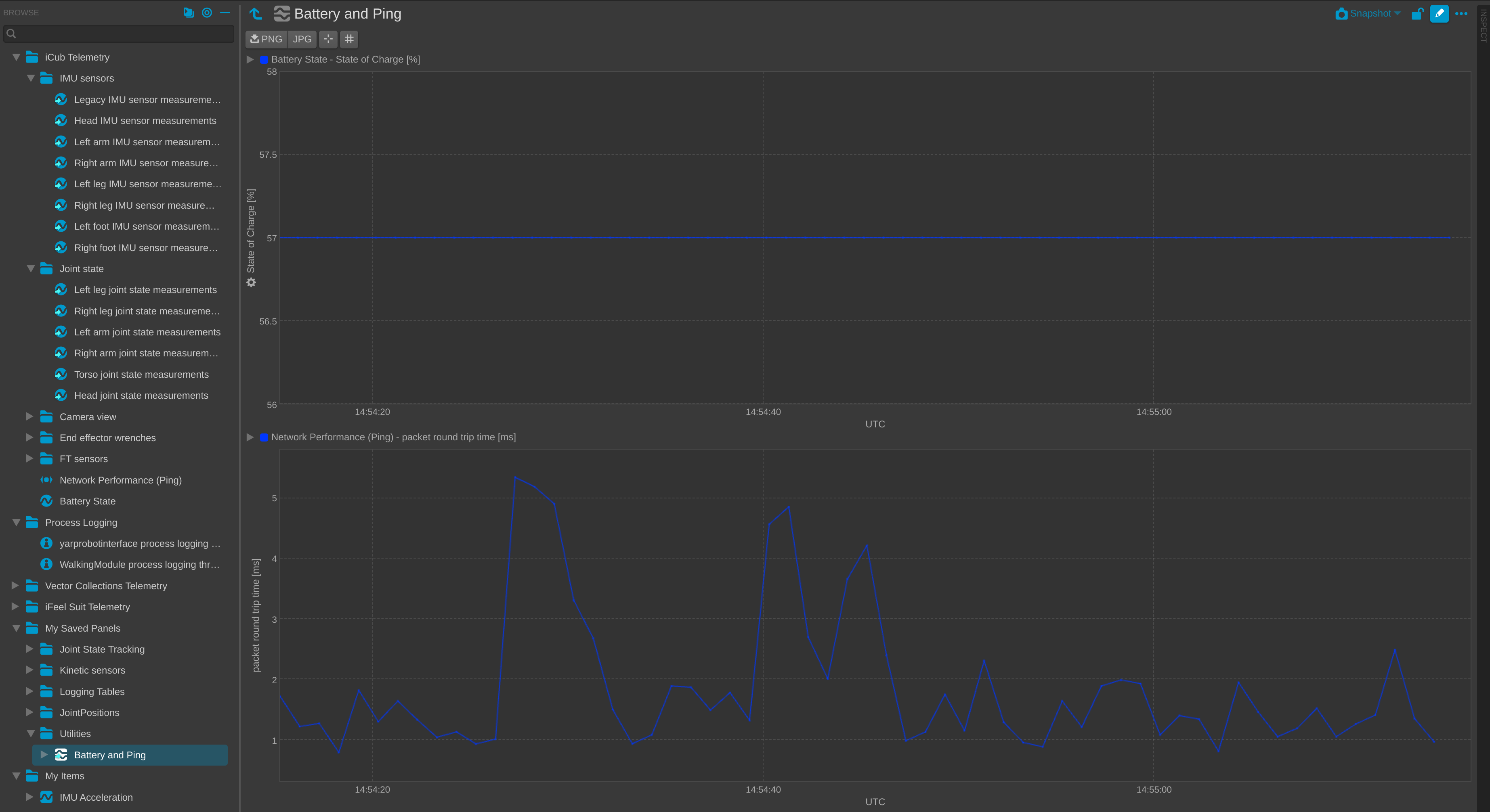}
              \put(5,45){\pgfsetfillopacity{0.5}\colorbox{black}{\pgfsetfillopacity{1.0}\textcolor{white}{\large{A}}}}
              \end{overpic}}\\[3pt]
    \subfloat{\begin{overpic}[width=\columnwidth]{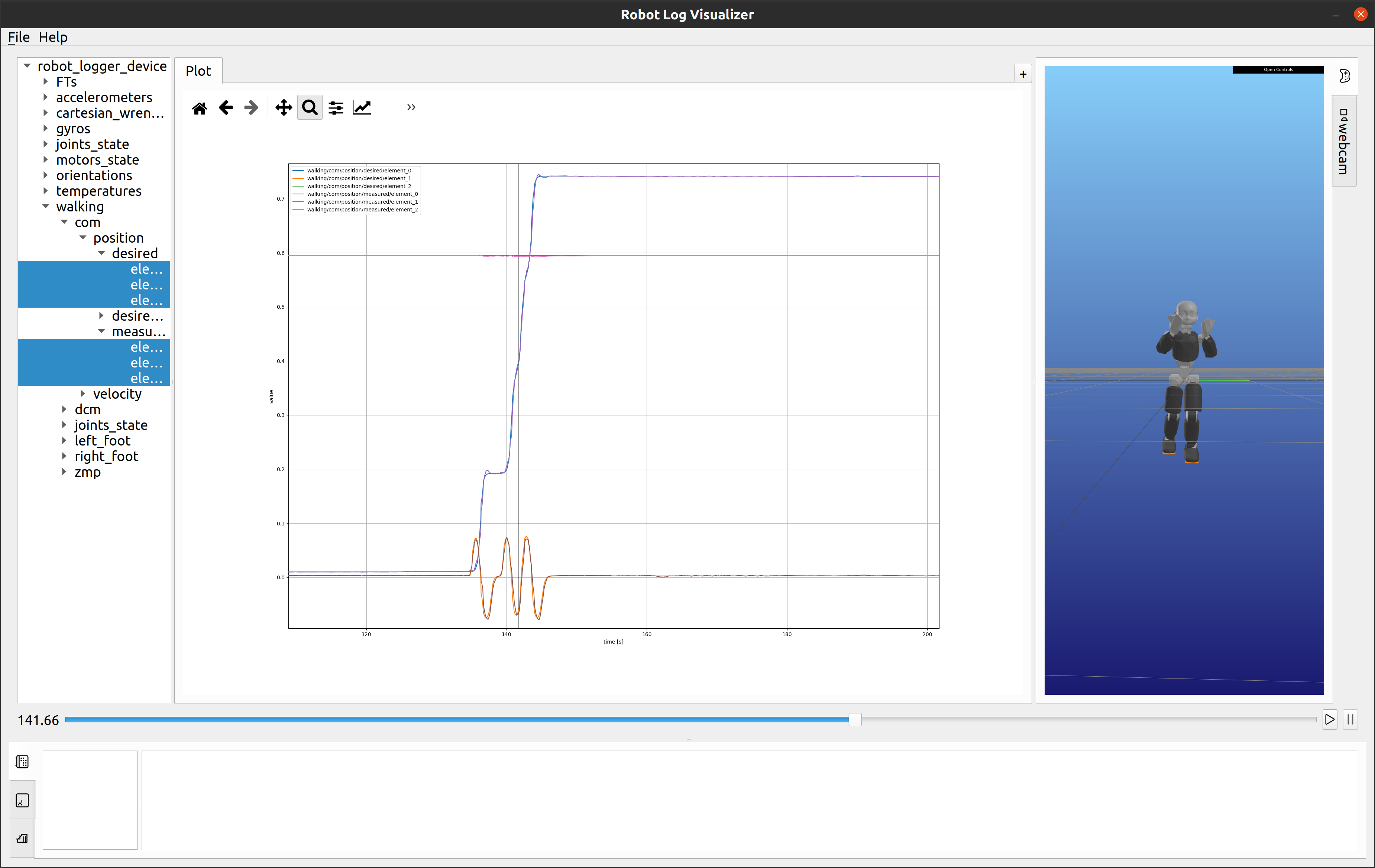}
              \put(5,55){\pgfsetfillopacity{0.5}\colorbox{white}{\pgfsetfillopacity{1.0}\textcolor{black}{\large{B}}}}
              \end{overpic}}
    \caption{\textbf{Logging systems}. Examples of usage of the logging systems presented in the ``Methods`` section. An example of the online logging system showing the battery level and the robot's head communication delay (\textbf{A}). The offline logging system displaying a representation of the robot and a plot (\textbf{B}).}
    \label{fig:Openmct}
\end{figure}

\end{document}